\pdfoutput=1

\documentclass[11pt]{article}

\usepackage[preprint]{acl}

\usepackage{times}
\usepackage{latexsym}
\usepackage{graphicx}
\graphicspath{{./images/}}
\usepackage{booktabs} 
\usepackage{color}  
\usepackage{amsmath}  
\usepackage{subcaption}
\usepackage{caption}
\usepackage{tikz}
\usepackage{colortbl} 
\usepackage{framed}
\usepackage{multirow}
\usepackage{multicol}
\usepackage{hyperref}
\usepackage{url}
\usepackage{balance}
\usepackage{verbatim}
\usepackage{cancel}
\usepackage{xspace} 
\usepackage[ruled,vlined]{algorithm2e}
\usepackage{bbold} 
\usepackage{balance}
\usepackage{stmaryrd}
\usepackage{enumitem}
\usepackage{array} 
\usepackage{tcolorbox} 
\usepackage{listings}  
\usepackage{tabularx}  

\usepackage[T1]{fontenc}

\usepackage[utf8]{inputenc}

\usepackage{microtype}

\usepackage{inconsolata}

\usepackage{graphicx}

\title{\mbox{Recursive Question Understanding for Complex Question Answering} \mbox{over Heterogeneous Personal Data}}

\author{Philipp Christmann\\
  Max Planck Institute for Informatics \\
  Saarland Informatics Campus \\
  \texttt{pchristm@mpi-inf.mpg.de} \\\And
  Gerhard Weikum \\
  Max Planck Institute for Informatics \\
  Saarland Informatics Campus \\
  \texttt{weikum@mpi-inf.mpg.de}}

\newcommand{\struct}[1]{\texttt{#1}}
\newcommand{\utterance}[1]{\textit{#1}}
\newcommand{\phrase}[1]{\textit{``#1''}}
\newcommand{\datakey}[1]{\texttt{``#1''}}

\newenvironment{Snugshade}[1][236,236,236]{
    \setlength{\itemsep}{0pt}
     \setlength{\parsep}{0pt}
     \setlength{\topsep}{0pt}
     \setlength{\partopsep}{0pt}
     \setlength{\leftmargin}{1.5em}
     \setlength{\labelwidth}{0em}
     \setlength{\labelsep}{0em} 
    \setlength{\parskip}{0pt}
    \definecolor{shadecolor}{RGB}{#1}%
    \begin{snugshade}
}{%
    \end{snugshade}%
}

\newcommand{\method}{\textsc{ReQAP}\xspace}
\newcommand{\benchmark}{\textsc{PerQA}\xspace}

\newcommand{\rag}{\textsc{Rag}\xspace}
\newcommand{\qg}{\textsc{CodeGen}\xspace}
\newcommand{\cg}{\textsc{CodeGen}\xspace}
\newcommand{\cog}{\textsc{CodeGen}\xspace}
\newcommand{\qud}[1]{\textit{\underline{#1}}\xspace}

\newcommand{\extract}{\texttt{EXTRACT}\xspace}
\newcommand{\retrieve}{\texttt{RETRIEVE}\xspace}
\newcommand{\operator}[1]{\texttt{\MakeUppercase{#1}}}

\newcommand{\squishlist}{
    \begin{list}{$\bullet$}{ 
        \setlength{\itemsep}{0pt}
        \setlength{\parsep}{1pt}
        \setlength{\topsep}{1pt}
        \setlength{\partopsep}{0pt}
        \setlength{\leftmargin}{1.5em}
        \setlength{\labelwidth}{1em}
        \setlength{\labelsep}{0.5em} 
    } 
}

\newcommand{\squishend}{
  \end{list}  }
  
\newcommand{\GW}[1]{{\color{blue}{GW: #1}} }

\newcommand{\myparagraph}[1]{\noindent \textbf{#1}.}

\newcommand{\relaxed}[1]{\small (${#1})$}
\newcommand{\relaxedb}[1]{\small ($\mathbf{{#1}}$)}

\newcommand{\qufunction}[1]{\textbf{\texttt{\MakeUppercase{#1}}}}
\newcommand{\quarg}[1]{{#1}}

\begin{document}

\maketitle
\begin{abstract}
Question answering over mixed sources, like text and tables,
has been advanced 
by verbalizing all contents and encoding it with a language model.
A prominent case of 
such heterogeneous data 
is personal information:
user
devices log vast amounts of data every day,
such as calendar entries, workout statistics,
shopping records, streaming history, and more.
Information needs range from simple look-ups to queries of analytical nature.
The challenge is to provide humans with convenient access with small footprint, so that all personal data stays on the user devices.
%
%
%
We present \method, a novel method that creates an executable
operator tree for a given
question, via recursive decomposition.
Operators are designed to enable seamless integration of structured and unstructured sources,
and the execution of the operator tree yields a traceable answer.
We further release the \benchmark benchmark, with persona-based
data and questions, covering a diverse spectrum of
realistic
user needs.
\end{abstract}


\section{Introduction}
\label{sec:intro}

\subsection{Motivation and Problem Statement}

Question answering (QA) over
combinations of structured and textual data has been substantially advanced
by verbalizing all contents and encoding it with a large language model (LLM)~\cite{DBLP:conf/acl/YinNYR20,DBLP:journals/pvldb/ThorneYSS0L21,DBLP:conf/naacl/OguzCKPOSGMY22}.

\begin{figure} [t]
     \includegraphics[width=\columnwidth]{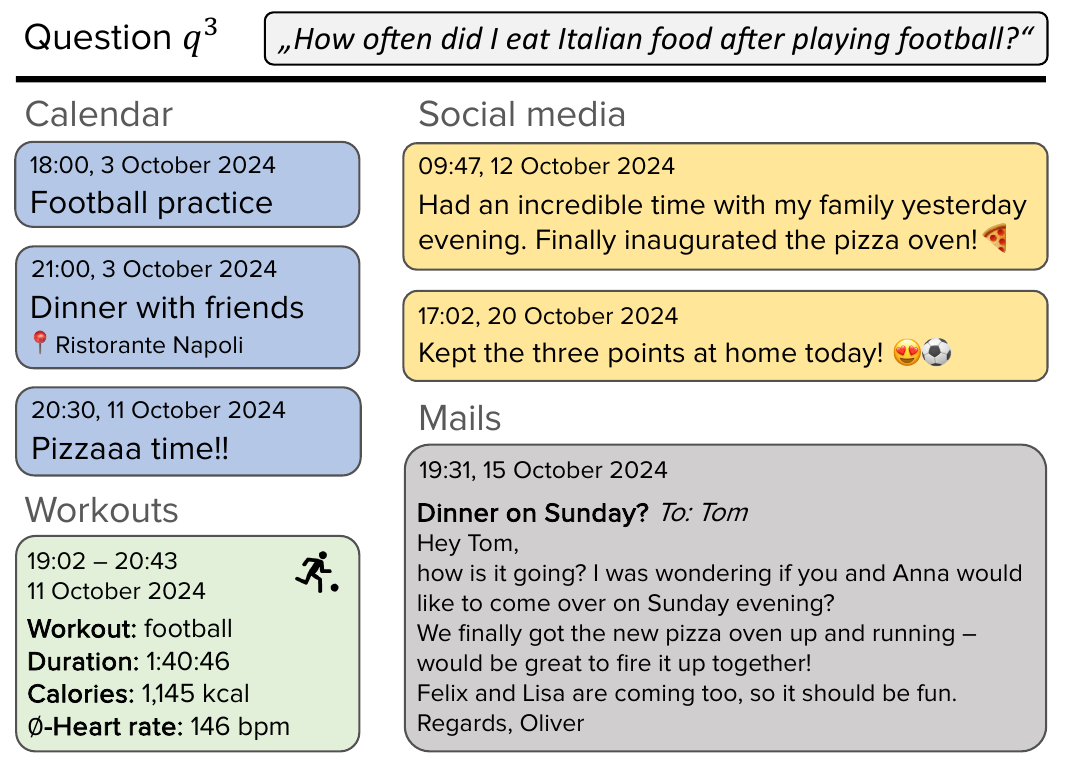}
     \vspace*{-0.6cm}
     \caption{Excerpt of relevant user data for the question $q^3$, stored in \textit{structured} (workouts), \textit{semi-structured} (calendar) and \textit{unstructured} (social media, mails) sources.}
     \label{fig:user-data}
     \vspace*{-0.3cm}
\end{figure}

A prominent case of such heterogeneous content, of growing importance, is {\em personal information management} \cite{halevy2024personal}.
User devices
collect
vast amounts of {personal data} 
on a daily basis.
Each interaction with digital services, such as Amazon Marketplace, Netflix or Spotify,
is stored for the specific user,
and services running on smartphones or smart watches track data without explicit user actions,
such as travel routes or workouts.
Users want convenient tools to query
their data,
as easy as with an LLM-powered AI co-pilot.
At the same time,
\textit{data protection} is a crucial requirement,
as full control over sensitive content is a top priority for users \cite{DBLP:conf/www/MansourSHZCGAB16,DBLP:conf/cikm/KalokyriBM18}.
Ideally, all data and its processing resides locally on the owner's devices (mobile, tablet or PC),
tightly limiting the available computational resources.
Reconciling these goals
is a challenge~\cite{halevy2024personal}.

Users' needs range from look-ups of events to queries of analytical nature \cite{DBLP:conf/acl/TanD0MSYH23}:
\begin{Snugshade}
\footnotesize
\noindent $q^1$: \utterance{Which books did I buy for my friends last Christmas?}\\
$q^2$: \utterance{When did I last have pizza with both my parents?}\\
$q^3$: \utterance{How often did I eat Italian food after playing football?}\\
$q^4$: \utterance{Daily kilometers I ran during business trips?}
\end{Snugshade}

Such questions require fine-grained understanding of the user intent, and entail complex operations like
\textit{grouping}, \textit{aggregations}, \textit{joins} or matching \textit{temporal conditions},
often with hundreds of relevant information pieces from different sources.

The goal of this work is to devise a QA system for personal data that
(i) supports such complex information needs,
(ii) is user-friendly and intuitive for lay people,
and 
(iii) has low footprint regarding computation, memory and energy,
suitable for running directly on user devices.

\vspace{-0.1cm}
\subsection{State of the Art and its Limitations}
%
\noindent{\bf Verbalization:}
State-of-the-art methods for QA over such heterogeneous data would mostly follow the \textit{verbalization} theme.
Relevant data is retrieved, linearized into textual form and cast into token sequences~\cite{DBLP:conf/naacl/OguzCKPOSGMY22, christmann2022conversational}.
These are fed into an LLM~\cite{minaee2024large} for 
generating answers,
in the spirit of {\bf retrieval-augmented generation
(\rag)}~\cite{DBLP:journals/pvldb/ThorneYSS0L21, liutapex, badaro2023transformers}.
%
%

Verbalization has been successfully applied for heterogeneous QA when few evidence pieces are sufficient for answering.
However, verbalization falls short 
as relevant user data often exceeds the context window of LLMs, going well beyond $100$K tokens (e.g., for $q^3$ and $q^4$).
Even if their context size was unlimited,
LLMs struggle with more complex operations over hundreds of events,
such as aggregation or grouping.
This limitation is amplified for numerical data,
which is encoded as simple tokens, hindering precise aggregation.

\vspace{0.1cm}
\noindent{\bf Translation:}
The predominant theme for such complex questions, with large-scale
aggregation and joins, is \textit{translation}.
An LLM
is provided with the natural language question and the data schema (i.e., headers of tables),
in order to generate code in the form of a logical query,
typically for languages like SQL or SPARQL (e.g.,~\citet{fan2024combining, DBLP:journals/pvldb/LiLCLT24,DBLP:journals/corr/abs-2408-05109}),
which is then directly executed.
This \textbf{code generation}
(\textbf{\cg})
is often referred to as NL2SQL or Text2SQL~\cite{DBLP:journals/pvldb/LiLCLT24,DBLP:journals/corr/abs-2408-05109}.

Translation works well when all data appears in a structured manner,
with a well-designed data schema, such that a single query can access all relevant information jointly.
In our setting, this is not the case, as exemplified in Figure~\ref{fig:user-data}.
Relevant user data naturally appears in text, which is not easily accessible by queries.
Processing such unstructured text is a key requirement in the context of QA over personal data (e.g., for $q^2$, $q^3$, and $q^4$).

\vspace{0.1cm}
Consider question $q^3$ as our running example, for which Figure~\ref{fig:user-data} shows a small excerpt of relevant user data.
Relevant
information, 
such as playing football,
could be stored in different forms,
including \textit{structured sources} like workout tables,
\textit{semi-structured sources} such as calendar entries,
or \textit{unstructured sources} such as social media posts.
Moreover, the very same event 
is often captured by multiple sources
(e.g., calendar entry on 11 October and social media post on 12 October), which could lead to double-counting.
Finally, cues are often subtle (e.g., the football icon in the second social media post),
which means that simple matching of surface text is not sufficient.

\begin{figure*} [tbh]
     \includegraphics[width=\textwidth]{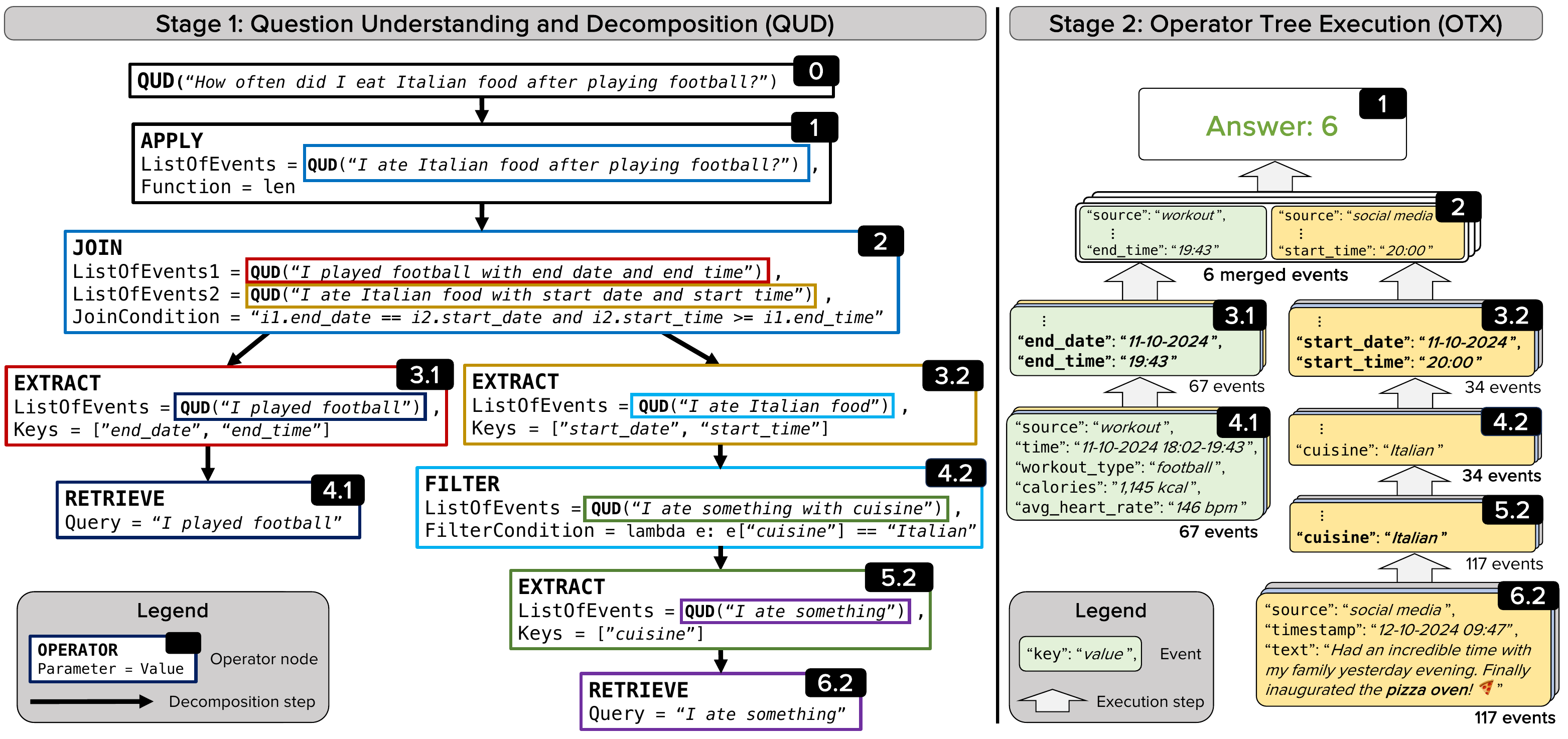}
     \vspace*{-0.6cm}
     \caption{Example inference of \method. The left side illustrates the output of the QUD stage, an operator tree, for the example question $q^3$. The right side shows how this tree is executed within the OTX stage to derive the answer.}
     \label{fig:overview}
     \vspace*{-0.2cm}
\end{figure*}

\vspace{-0.1cm}
\subsection{Approach and Contributions}

To overcome these limitations, the key idea of this work is to bridge the gap between verbalization and translation,
and combine the strengths of both paradigms.
Our approach creates an \textit{operator tree}, similar to a query in the translation paradigm,
for which we propose judiciously designed operators that enable 
access
to unstructured text, based on ideas of the verbalization paradigm.

Specifically, we present a novel methodology for the task of QA over personal data,
\method (\underline{Re}cursive \underline{Q}uestion Understanding for Complex Question \underline{A}nswering over \underline{P}ersonal Data).
All personal data sources are treated as lists of events described by 
key-value pairs.
For a given user question, \method derives a database-style operator tree.
Notably, in order to generate these operator trees, we propose a new mechanism for recursively
\textit{decomposing} the complexity of
questions,
via in-context learning and model distillation of small-scale LMs.
This enables deployment of the entire QA system on end user devices.
Apart from standard SQL-like operators, we introduce two novel operators,
\retrieve and \extract,
that enable efficient retrieval from both structured and unstructured data sources,
and enhancing unstructured text with question-related key-value pairs,
via on-the-fly information extraction.
These operator trees are executed for deriving the traceable answer.

For evaluation, we construct a new benchmark \benchmark, which synthesizes realistic user data and questions based on personas,
and has $3{,}500$ complex questions and more than $40{,}000$ events per persona. 
Experiments on \benchmark demonstrate substantial
improvements of \method compared to
verbalization or translation
baselines.

\vspace*{0.1cm}
Salient contributions of this work are\footnote{Code and data available at \href{https://reqap.mpi-inf.mpg.de}{https://reqap.mpi-inf.mpg.de}.}:
\squishlist
\item 
    a novel approach, \method, for QA over heterogeneous personal data, based on a recursive question understanding mechanism;
\item 
    a complete QA system that can run on end user devices,
    trained via LLMs and their in-context learning capabilities;
\item 
    a new large-scale benchmark for personal QA, \benchmark, with realistic data and questions.
\squishend

\vspace*{-0.2cm}
\section{The \method Data Model}
\label{sec:datamodel}
We treat all data sources as
temporally ordered lists
of {\bf events},
where each event is a dictionary with a set of {\bf key-value pairs}.
For example, each mail is an event with keys \struct{date}, \struct{time}, \struct{sender}, \struct{recipients}, \struct{subject} and \struct{body};
for music streams, each event would have keys like \struct{date}, \struct{start\_time}, \struct{end\_time}, \struct{title}, \struct{artists}.
Example events are shown in the bottom right of Figure~\ref{fig:overview} (workout in light green and social media in yellow).

Our current implementation covers the following event types,
accessed via data exports:
\textit{calendar entries}, \textit{notes}, \textit{mails} and \textit{social media posts} with semi-structured and unstructured data,
and \textit{workouts},
streams (\textit{movies}, \textit{TV series}, \textit{music}),
and \textit{online purchases} with structured data.
Note that event types can be added easily with this agile data model, which treats all contents as key-value pairs.

\vspace*{-0.1cm}
\section{The \method Methodology}
\label{sec:method}

\vspace*{-0.1cm}
\method
processes questions in two stages:\\
\noindent (i) the {\bf question understanding and decomposition (QUD)} for constructing the operator tree,
and (ii) the {\bf operator tree execution (OTX)} stage for deriving the answer with the corresponding events.
Figure~\ref{fig:overview}
shows a characteristic example 
that illustrates the two stages of \method.

\vspace*{-0.2cm}
\subsection{Question Understanding\&Decomposition}
\label{sec:QUD}



\myparagraph{Inference via ICL}
Due to the absence of appropriate training data,
we leverage 
\textit{in-context learning} (ICL)~\cite{brown2020language}
to generate our operator trees.
We construct handcrafted operator trees
for a diverse set of $40$ questions,
and then \textit{dynamically} retrieve $8$ ICL examples at each step via BM25.

In principle, we could prompt the LLM to generate a complete 
operator tree
in a single shot,
following the 
code generation line of works~\cite{DBLP:journals/pvldb/LiLCLT24}.
However, this approach is often too complex, even for an LLM,
with many output trees being incomplete or wrong.
Instead, we devise a novel method for {\em recursive decomposition}, where we call the LLM repeatedly to generate 
{\em partial operator trees} that have the freedom to resolve sub-questions of the user's input in \textit{recursive} QUD calls. 
These sub-questions are then the input to subsequent LLM calls, in order to refine the partial operator tree.
The recursion stops when
all leaf nodes are properly mapped to operators without any open
QUD calls.

\vspace*{0.1cm}
\noindent {\bf Example:}
Consider the operator tree in Figure \ref{fig:overview}.\\
\noindent
(i) The first LLM call generates node 1, with an {\em aggregate function} (list length, i.e., counting) applied to the question results.\\
\noindent
(ii) This is refined into node 2, with the {\em join} over two {\em sub-questions}:  \phrase{I played football} and \phrase{I ate Italian food}.\\
\noindent
(iii) Each of two sub-questions is translated into an {\extract} operator, 3.1 and 3.2, and the left branch invokes the leaf-node {\retrieve} operator 4.1.\\
\noindent
(iv) The right branch 3.2 is further refined, leading to a cascade of \operator{Filter} (4.2), \extract (5.2) and \retrieve (6.2) operators.\\
\noindent
(v) This operator tree is then executed bottom-up, as visualized on the right side of Figure~\ref{fig:overview}.

To facilitate this method,
our
ICL examples for the
LLM follow the same approach,
showing how a complex question can be split
into an operator part and a simpler sub-question.
Thus, each
few-shot sample
is a sequence of decomposition steps.

%
%


\vspace{0.1cm}
\myparagraph{Training}
%
Recall that we aim to deploy \method on the users' personal devices directly, including mobiles.
This would be impossible with the
large models ($\ge$$70$B parameters)
used above for ICL.

The solution is to harness the (question, operator-tree) pairs obtained 
via ICL to
\textit{distill}~\cite{hsieh2023distilling} the LLM into a smaller model (e.g., LLaMA with $1$B parameters).
To ensure high-quality training samples, we 
generate multiple operator trees for each training question,
evaluate them on the data, and keep only those
that return correct results.
%

\vspace*{0.1cm}
\myparagraph{Inference via SFT}
These \textit{supervised fine-tuned} (SFT) models are deployed
for generating operator trees
from user questions,
and follow the same decomposition
procedure as
with ICL.
In our experiments, we report on different QUD model sizes.

\subsection{Operator Tree Execution} 
The OTX stage executes the operator tree
obtained in the QUD stage to derive the answer.
Table~\ref{tab:operators} provides the full list of operators.
The following describes the two most important ones:
\retrieve and \extract.
Other operators and support functions 
(e.g., \operator{Apply} for invoking arbitrary Python functions,
or \operator{Unnest} for enumerating elements in a list), are
defined in Appendix \ref{app:operators}.

\begin{table} [t] 
\centering
\resizebox*{0.9\columnwidth}{!}{
\setlength{\tabcolsep}{2pt}
        \begin{tabular}{l}
          \toprule
                \qufunction{retrieve} (\quarg{Query, \textcolor{gray}{Optional: ListOfEvents}})  \\
                \qufunction{extract} (\quarg{ListOfEvents, Keys}) \\
            \midrule 
                \qufunction{join}  (\quarg{ListOfEvents1, ListOfEvents2, JoinCondition}) \\ 
                \qufunction{group\_by}  (\quarg{ListOfEvents, ListOfKeys}) \\
            \midrule
                \qufunction{filter}  (\quarg{ListOfEvents, FilterCondition}) \\
                \multicolumn{1}{l}{\qufunction{map} (\quarg{ListOfEvents, Function}) }  \\
                \multicolumn{1}{l}{\qufunction{apply} (\quarg{ListOfEvents, Function})} \\

                \multicolumn{1}{l}{\qufunction{unnest} (\quarg{ListOfEvents, KeyToUnnest}) } \\
           \midrule
                {\qufunction{argmin}, \qufunction{argmax}, \qufunction{min}, 
                \qufunction{max}, \qufunction{sum}, \qufunction{avg}} \\
            \bottomrule
        \end{tabular}
}
    \vspace*{-0.3cm}
    \caption{The \method operators with their input types.}
    \label{tab:operators}
    \vspace*{-0.2cm}
\end{table}

\subsubsection{\retrieve Operator}
\label{sec:retrieve}
\noindent{\bf Functionality.} 
The \retrieve operator 
takes as input a
\textit{query} generated by the QUD stage.
The output consists of \textit{all} events
from the personal data sources
that soft-match the
query,
via
matching against all key-value pairs.
%

\vspace{0.1cm}
\noindent{\bf Challenges.}
Unlike standard \rag settings, the \retrieve operator faces
specific challenges for analytic queries.
Many of these aggregate over hundreds of relevant events, and missing a
fraction of these would already distort the final answer.
Therefore, we need to ensure that this operator has {\em high recall},
ideally near 100\%, without 
adding too many false positives.
Moreover, the qualifying events identified by \retrieve may have redundancy,
as the same event can be mentioned in calendar, workout, social media, email etc. 
Therefore, de-duplication of redundant entries is crucial.

\vspace{0.1cm}
\noindent{\bf Execution.}
The \retrieve operator conceptually scans {\em all} personal data sources 
and can find matches among all their key-value pairs.
However, our method aims to reduce the computational costs by early pruning of
entire sources that do not contribute. 
This goal is achieved by executing the operator in a pipeline with five steps:

\noindent
{\bf Step 1:} We use the efficient {\em sparse
retriever SPLADE} \cite{DBLP:conf/sigir/FormalLPC22} to obtain an initial pool of candidate events, aiming at high recall (i.e., minimizing the risk of missing a good result).
All events with a non-zero score >$0.1$ are retained.
%

\noindent
{\bf Step 2:} The resulting pool of events is analyzed to identify {\em patterns}:
key-value pairs
that occur frequently (i.e., above a
specified
frequency threshold) in the candidate pool;
for example, \struct{subject}=\phrase{Team meeting} in calendar or
\struct{type}=\phrase{movie}.
In addition, for each source
with matching events,
we add a specific pattern (e.g., \struct{source}=\phrase{mail}),
to score
the entire source at once.

\noindent
{\bf Step 3:} Finally, the patterns, each representing a set of
events,
are classified into three categories:
(a) relevant (all events are kept), or
(b) completely irrelevant (all events are dropped),
or (c) partially relevant (events further filtered by key-value pairs).

\noindent{\bf Step 4:} 
A second classifier runs on all remaining events in (c). 
It labels key-value pairs into two bins:
(a) keep (for downstream operators) or (b) drop.

\noindent{\bf Step 5:} 
Finally, to tame the potential redundancy of qualifying events (e.g., a calendar entry and a mail on the same meeting),
we perform de-duplication. Whenever two events have overlapping time intervals, we combine them into a single event with all key-value pairs and the union of the timespans.

\vspace{0.1cm}
The pruning of entire sources in step 3 (e.g., music streams for queries on food or friends) yields a massive speed-up, and early elimination of frequent key-value pairs also contributes to the efficiency of \method
(e.g.,  disregarding all workouts with \struct{type}=\phrase{gym} for queries on football).
At the same time, to ensure high recall, 
our pipeline is
designed to
avoid unduly dismissing relevant events.

\vspace*{0.1cm}
\noindent {\bf Training.}
For steps 3 and 4, the two classifiers are implemented as 
{\em cross-encoders} (CE)~\cite{DBLP:journals/corr/abs-2403-10407,DBLP:series/synthesis/2021LinNY}
for high precision.
Our experiments use a standard CE\footnote{\href{https://huggingface.co/cross-encoder/ms-marco-MiniLM-L-12-v2}{cross-encoder/ms-marco-MiniLM-L-12-v2}},
fine-tuned with examples for pattern and event classification.
%

We derive training data from our dataset:
for each question, we identify relevant events as the ones used by the ground-truth query (as present in the \benchmark dataset).
For each \retrieve call in the operator tree, we would then add these events as positive training instances,
and uniformly sample hard negatives from events retrieved by SPLADE.
Patterns are labeled based on all matching events, as either
(a) fully relevant, (b) fully irrelevant, or (c) partially relevant.


\subsubsection{\extract Operator}
\noindent{\bf Functionality.}
The \extract operator 
takes as input
a set of {\em keys}, like attributes/field names, and a list of {\em events}, typically returned by an upstream \retrieve operator (i.e., child node in the operator tree).
The operator generates values for all combinations of events and keys in the input,
and augments events with these new key-value pairs.

\vspace{0.1cm}
\noindent{\bf Challenges.}
The \extract operator faces some very difficult cases when the input key does not semantically match any of the data keys
(beyond simple cases such as matching \phrase{day} to \struct{date} in calendar or workout data).
The challenges arise when the
operator's input key
requires \textit{extracting} information from a longer text field,
such as the body of a mail or the content of a social media post.
Example keys would be
\phrase{participants},
\phrase{trip destination},
or \phrase{cuisine} -- none of which equals any existing data key.
Note that the number of input events can often be in the order of hundreds or thousands, necessitating an efficient implementation.

\vspace{0.1cm}
\noindent{\bf Execution.}
%
Hence, we
utilize a
small-scale \textit{sequence-to-sequence model},
which is given
(i) the input key,
(ii) the \textit{verbalized} key-value pairs, and
(iii) notable user information (e.g., a list of friends) as input,
and generates a short phrase as output.
For the running example about Italian food after football
(question $q^3$ in Figure~\ref{fig:user-data}),
the input key is \phrase{cuisine}
and the value \phrase{Italian} is generated based
on the phrase \phrase{...pizza oven...} in the mail body.
For efficiency,
\extract 
creates
a frozen mapping
for simpler input keys (e.g., \phrase{date}$\to$\struct{start\_date}),
covering cases where at least 70\% of the first 50 inputs
lead to the very same
event-key (\struct{start\_date}).
For the remaining input events,
this mapping is then applied directly, avoiding further generation.

%
\vspace{0.1cm}
\noindent {\bf Training.}
To enable this rich behavior, we fine-tune a small LM, 
like BART~\cite{lewis2020bart}.

\begin{figure} [t]
     \includegraphics[width=\columnwidth]{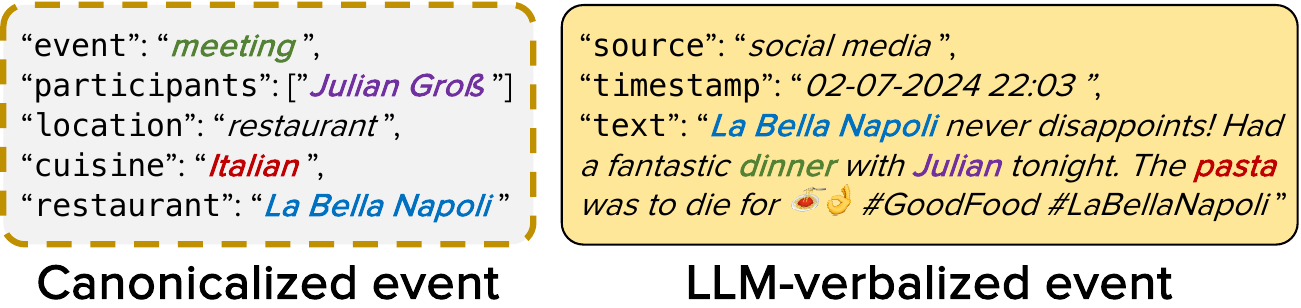}
     \vspace*{-0.6cm}
     \caption{Example canonicalized event (not accessible at inference), with the corresponding verbalized version.}
     \label{fig:events}
     \vspace*{-0.2cm}
\end{figure}

We derive high-quality training data from our dataset,
which provides parallel data of canonicalized events and their verbalized counter-parts.
Figure~\ref{fig:events} shows an example
of a canonicalized entry representing a meeting,
and a corresponding social media post that is generated via an LLM.

\section{The \benchmark Dataset}
\label{sec:dataset}

\subsection{Prior Benchmarks}
There is a wealth of established benchmarks for all kinds of
QA~\cite{DBLP:journals/csur/RogersGA23}.
However, almost all of these are centered on objective contents about real-world facts, and thus unsuitable for our setting. QA over personal data needs to consider subjective events including personally sensitive details.
Such data is subject to privacy laws, 
and this explains the lack of an appropriate benchmark.

To the best of our knowledge, the only existing data source that is suitable for our task is TimelineQA \cite{DBLP:conf/acl/TanD0MSYH23},
which
provides code for generating persona-based contents and questions.
It allows generating personal data at large scale.
However, the specified attributes for synthetic personas are basic, and user content is generated based on 
a limited number of simple templates.
Further, complex questions are derived from solely $42$
templates (vs. $\approx$2,000 in \benchmark).
Thus,
TimelineQA exhibits very limited diversity,
not suitable to capture real-world challenges in personal QA.

\subsection{Benchmark Creation}

To bridge this gap in the existing literature,
we devise a principled mechanism to create
the \benchmark dataset.
The construction proceeds in three steps:
(i) \textit{persona collection}, in which we let humans (co-workers at our research group) craft realistic and detailed biographical information on \textbf{fictional persons},
(ii) \textit{user data creation}, which makes use of the persona information (such as their music or travel preferences) and creates large amounts of realistic user data,
and (iii) \textit{question generation}, that generates complex questions, and their ground-truth answers, for the personas and their user data.

\vspace{0.1cm}
\myparagraph{Persona collection}
The personas are created by having humans (from our research group) complete questionnaires with $30$ fields about age, gender, family, education, work history, traveling, hobbies, fitness activities, interests in music and movies, food tastes and other preferences. 
We collect $20$ such human-crafted personas.

\begin{table} [tb]
    \centering
\resizebox*{0.95\columnwidth}{!}{
    \setlength{\tabcolsep}{3pt}
    \renewcommand{\arraystretch}{0.8}
    \newcolumntype{G}{>{\columncolor [gray] {0.90}}c}
    \begin{tabular}{l G c G}
        \toprule
            \textbf{Category $\downarrow$ / Split $\rightarrow$} & \textbf{Train} & \textbf{Dev} & \textbf{Test} \\
        \midrule
            \textbf{Personas}    & $12$ & $2$ & $6$ \\
            \textbf{Question templates}    & $1{,}200$ & $170$ & $600$ \\
            \textbf{Unique questions}    & $2{,}571$ & $170$ & $826$ \\
        \midrule
            \textbf{Movie streams}    & $1{,}257$  & $424$  & $1{,}245$  \\
            \textbf{TV series streams}    & $238$  & $148$  & $230$  \\
            \textbf{Music streams}    & $33{,}754$  & $31{,}616$  & $38{,}257$  \\
            \textbf{Workouts}    & $635$  & $622$  & $822$  \\
            \textbf{Online purchases}    & $531$  & $1{,}074$  & $850$  \\
        \midrule
            \textbf{Calendar entries}    & $1{,}552$  & $1{,}367$  & $2{,}106$  \\
            \textbf{Mails}    & $506$  & $490$  & $636$  \\
            \textbf{Social media posts}    & $670$  & $579$  & $819$  \\
        \midrule
            \textbf{Total events (per persona)}    & $39{,}142$  & $36{,}322$  & $44{,}965$  \\
        \bottomrule
    \end{tabular} 
}   
    \vspace*{-0.3cm}
    \caption{Statistics for the \benchmark dataset.}
    \label{tab:benchmark-in-paper}
    \vspace*{-0.4cm}
\end{table}

\vspace*{0.1cm}
\myparagraph{Canonicalized events}
The next step is to create large-scale
event repositories for each persona.
In this stage, events are \textit{canonicalized}:
for each event type, events have a specific set of key-value pairs.
This allows us to derive ground-truth answers later.

For streaming data (movies, TV series, music), we utilize Wikidata~\cite{vrandevcic2014wikidata}
to derive real-life entities matching the persona preferences (e.g., the favorite genre and songs).
Similarly, for online shopping, we utilize a public data dump~\cite{ni2019justifying}.
For realistic travel itineraries in-line with the persona information,
we prompt GPT4o.
For workouts, we utilize public data from Endomondo \cite{ni2019modeling}.


We further add special events, like personal milestones
(marriage, birth of a child, new job,\dots)
based on the personas.
Finally, we add regular events like doctor appointments,
meetings with friends, or New Year's parties.
This is done based on pre-defined probability values,
inspired by the procedure in \citet{DBLP:conf/acl/TanD0MSYH23}.

On average, for each persona, we create $40$K such canonicalized events.
The left side of Figure~\ref{fig:events} provides an example canonicalized event.
A list of all event types is provided in Appendix~\ref{app:benchmark}.

\vspace{0.1cm}
\myparagraph{Observable events}
In reality, events such as restaurant visits, birthday parties, or a child birth, are usually not captured in a canonicalized form, but in unstructured text as in calendar entries, mails and social media posts.
We utilize LLaMA3.1\footnote{\label{llama31}\href{https://huggingface.co/meta-llama/Llama-3.1-70B-Instruct}{meta-llama/Llama-3.1-70B-Instruct}}
to verbalize such events,
and create realistic and diverse user data.
Figure~\ref{fig:events} illustrates the connection between canonicalized and verbalized events.
Streams, workouts and online purchases are kept in a canonicalized form with high probability,
but are also verbalized
with a relatively small probability.


\vspace{0.1cm}
\myparagraph{Generation of complex questions}
For generating questions, we handcraft a diverse set of $40$ complex questions to start with
-- the same questions that are used for ICL in the QUD stage.
We then iteratively sample up to $20$ questions from this set,
and prompt an LLM (with a temperature of $0.6$) to generate $25$ new questions.
In addition to the questions, the ICL examples (and the model outputs) also contain
SQL queries,
run on a database with the canonicalized events,
to derive the ground-truth answers.
Figure~\ref{fig:prompt-questions} shows the instruction used for creating these question-query pairs.
We utilize GPT4o and LLaMA3.1\textsuperscript{\ref{llama31}} to construct
$1{,}000$ questions each, for enhanced diversity.
Generated question-query pairs are de-duplicated,
and checked for correctness semi-automatically.

\vspace{0.1cm}
\myparagraph{Deriving \benchmark}
We release \benchmark with $20$ personas, $40$K events per persona, and $3{,}567$ unique questions.
This data is split along two dimensions,
personas and question templates,
to derive the
train ($12$ personas, $1{,}200$ templates),
dev ($2$ personas, $170$ templates),
and test ($6$ personas, $600$ templates) sets.
Question templates are initiated based on the
personas,
to ensure natural and meaningful questions. Table~\ref{tab:benchmark-in-paper} shows statistics
of \benchmark,
and further details are provided in Appendix~\ref{app:benchmark},
including example questions and data in 
Tables~\ref{tab:perqa-questions} and~\ref{tab:perqa-data}.



\section{Experimental Setup}
\label{sec:setup}

\myparagraph{Configurations}
We experiment with three different options for QUD:
using \textbf{GPT4o} ({\small ``{gpt-4o}''}) or \textbf{LLaMA3.3}\footnote{\href{https://huggingface.co/meta-llama/Llama-3.3-70B-Instruct}{meta-llama/Llama-3.3-70B-Instruct}}
with $8$
ICL examples,
or a $1$B model\footnote{\href{https://huggingface.co/meta-llama/Llama-3.2-1B-Instruct}{meta-llama/Llama-3.2-1B-Instruct}} 
fine-tuned as described in Section~\ref{sec:QUD} (\textbf{SFT}).
%
To initialize the \extract operator, we use BART-base\footnote{\href{https://huggingface.co/facebook/bart-base}{facebook/bart-base}},
and the \retrieve operator uses
SPLADE\footnote{\href{https://huggingface.co/naver/splade-cocondenser-ensembledistil}{naver/splade-cocondenser-ensembledistil}} and a cross-encoder pre-trained on MSMARCO\footnote{\href{https://huggingface.co/cross-encoder/ms-marco-MiniLM-L-12-v2}{cross-encoder/ms-marco-MiniLM-L-12-v2}}.
%
Further details
for model training
are
in Appendix~\ref{app:setup}.

\vspace{0.1cm}
\myparagraph{Baselines}
We compare
\method
against
\rag and \qg:


\squishlist
\item 
\noindent{\bf \rag:}
This retrieval-augmented generation follows the \textit{verbalization} paradigm.
For retrieving relevant events, a two-stage technique is employed, with 
SPLADE retrieval followed by a fine-tuned cross-encoder (similar as in~\ref{sec:retrieve}).
\item 
\noindent{\bf \qg:}
We also compare against a code generation baseline following the \textit{translation} paradigm.
The input is the user question and the SQL schema of the personal data,
and the output is a SQL query.
For fine-tuning, we utilize SQL queries that lead to the correct answer
(similar as in~\ref{sec:QUD}).
\squishend

For fair comparison, these baselines use the same model variants as the QUD stage of \method:
ICL with $8$ examples using {GPT4o} or {LLaMA3.3},
or a fine-tuned
decoder-only
LM.


\vspace*{0.1cm}
\myparagraph{Metrics}
We measure 
\textbf{Hit@1} as our main metric
for personal QA, where users expect a single concise answer.
As some questions aggregate over hundreds of events,
we also measure 
a relaxed metric \textbf{Rlx-Hit@1}
which allows a $\pm 10$\% slack for numeric answers.
Both metrics are micro-averaged over all questions in the test set.
Statistical significance of improvements by \method
compared to the best baseline (for each model) is measured via McNemar's test with $p$<$0.05$,
and is indicated by an asterisk(*).
\section{Experimental Results}
\label{sec:experimental-results}


\subsection{Key Findings}
\label{sec:main-results}



\myparagraph{\method clearly wins over baselines}
Table~\ref{tab:main-res} shows the main results on \benchmark.
First of all, \method substantially 
outperforms baselines
for each variant.
The difference is most pronounced with LLaMA3.3, where \method improves
Hit@1 by $8.3$ percentage points compared to \cog, the best baseline.
The same observation holds for Rlx-Hit@1, 
with similar improvements over baselines.

\vspace{0.1cm}
\myparagraph{\rag fails on \benchmark}
While the baselines based on \cog remain competitive,
\rag performs poorly.
It cannot cope with questions that 
need to aggregate (sum, avg, ordering, etc.)
over hundreds of events.
In some cases, the
input data exceeds the maximum context size of the LM ($128$K tokens).

\vspace{0.1cm}
\myparagraph{SFT is a viable solution for on-device inference}
The variants based on GPT4o and LLaMA3.3 are useful for comparison purposes,
but cannot be run directly on user devices.
Remarkably, our SFT variant, which is suitable for
on-device inference,
exhibits competitive performance even with a 100x smaller model size.
Its Hit@1 of 0.380 is almost as good as the best observed score of 0.386 (by \method with GPT-4o), and its relaxed Hit@1 of 0.53 is the best across the board.

\vspace{0.1cm}
\myparagraph{Approximate answers are good} 
The questions in \benchmark are difficult for all methods, with Hit@1 being below $40$\%.
Nevertheless, \method often yields decent order-of-magnitude answers.
An indicator is the Rlx-Hit@1 metric, for which \method achieves 
53\% with the small SFT variant.
Recall that this allows $\pm 10\%$ slack, which is on the conservative side. 
Even more relaxed approximations could still be tolerable by users.

\begin{table} [t]
    \small
    \centering
    \newcolumntype{G}{>{\columncolor [gray] {0.90}}c}

    \resizebox*{\columnwidth}{!}{
    \begin{tabular}{l G c G}
        \toprule
            \textbf{Model $\rightarrow$} & \textbf{GPT4o} & \textbf{LLaMA3.3}  & \textbf{SFT}\\
            \textbf{Method $\downarrow$}  &  ($\gg$100B) &  (70B) & (1B) \\
        \midrule
            \multirow{2}{*}{\textbf{\rag}}    & ${0.149}$  & $0.123$  & $0.029$  \\
                 & \relaxed{0.20} & \relaxed{0.18}  &  \relaxed{0.06} \\

        \midrule
            \multirow{2}{*}{\textbf{\qg}}  & $0.319$ & $0.239$& $0.315$  \\
             &  \relaxed{0.44}  & \relaxed{0.33} & \relaxed{0.47} \\

        \midrule
            \multirow{2}{*}{\textbf{\method} (ours)}    & $\mathbf{0.386}$*  & $\mathbf{0.322}$*   & $\mathbf{0.380}$*   \\
               & \relaxedb{0.52}*  &  \relaxedb{0.46}*  &  \relaxedb{0.53}*  \\
        \bottomrule
    \end{tabular}
    }
    \vspace*{-0.3cm}
    \caption{Main results on \benchmark \textbf{\textit{test}} set. Metric is Hit@1, and Rlx-Hit@1 in parentheses.}
    \label{tab:main-res}
    \vspace*{-0.1cm}
\end{table}

\subsection{Analysis}
\label{sec:analysis}

\begin{table}[t]
{
    {
        \setlength{\tabcolsep}{2.5pt}
        \footnotesize
        \centering
        \newcolumntype{G}{>{\columncolor [gray] {0.90}}c}
        \begin{tabular}{l c G G G}
            \toprule
                \textbf{Complexity} & \textbf{No. of} & \textbf{\rag} & \textbf{\cog} & \textbf{\method} \\
                 & \textbf{questions} & {(GPT4o)} & {(GPT4o)} & {(GPT4o)} \\
            \midrule
                \textbf{Ordering}     & $1{,}236$ & $0.167$ & $0.440$ & $\mathbf{0.529}$ \\
                \textbf{Grouping}     & $931$  & $0.172$ & $0.444$ & $\mathbf{0.537}$ \\
                \textbf{Temporal}     & $1{,}945$ & $0.129$ & $0.290$ & $\mathbf{0.417}$ \\
                \textbf{Aggregation}  & $1{,}790$ & $0.130$ & $0.228$ & $\mathbf{0.296}$ \\
                \textbf{Join}         & $369$  & $0.073$ & $0.176$ & $\mathbf{0.236}$ \\
                \textbf{Multi-source} & $1{,}710$ & $0.196$ & $0.237$ & $\mathbf{0.365}$ \\
            \bottomrule
        \end{tabular}
    }
    \vspace*{-0.3cm}
    \caption{Performance by complexity types of questions.}
    \label{tab:category-performance}
    \vspace*{-0.3cm}
}
\end{table}

\vspace{-0.1cm}
\myparagraph{\method performs well across question types} 
We categorized questions in \benchmark into different complexity types:
{ordering} (e.g., {\small \phrase{Which song did I listen to the most during my last trip?}}),
{grouping} ({\small \phrase{which four restaurants do I go to the most?}}),
{temporal} ({\small \phrase{How many times did I visit the doctor in the last 3 years?}}),
{aggregation} ({\small \phrase{What was the total amount spent on online purchases last year?}}),
{join} ({\small \phrase{How many times did I watch a movie after a run?}}), and
{multi-source} ({\small \phrase{which of my trips was the shortest?}}).
Note that each question can be assigned multiple such complexity types.

\method consistently outperforms
baselines,
with the most substantial gains for temporal questions.
\method performs especially well on questions requiring ordering and grouping, while join-questions are the most challenging for all methods.

\vspace{0.1cm}
\myparagraph{\retrieve and \extract operators are decisive} 
Table~\ref{tab:ablation} shows
ablation studies for \method,
when
(i) the operator tree is generated in one shot, \textit{without recursive decomposition},
(ii) all events retrieved by SPLADE are retained,
\textit{without cross-encoders},
(iii) values are extracted from events 
only by exact matches of keys,
\textit{without LM generation}.

Performance degrades dramatically for each of these variants.
Most notably,
the devised implementation of the \retrieve and \extract operators
substantially contribute to \method's performance.
Also, the recursive decomposition of the question intent
strikes as a key asset
as
generating the operator trees in a single step is prone to errors.

\begin{table} [t]
{
    \footnotesize
    \centering
    \newcolumntype{G}{>{\columncolor [gray] {0.90}}c}
    \begin{tabular}{l G G}
        \toprule
            \textbf{Method $\downarrow$ / Metrics $\rightarrow$} & \textbf{Hit@1} & \textbf{Rlx-Hit@1} \\
        \midrule
            \textbf{\method} (SFT)    & $\mathbf{0.396}$ & $\mathbf{0.54}$  \\
        \midrule
            \textbf{w/ QUD\,-- one-shot}    & $0.356$ & $0.50$  \\
            \textbf{w/ \retrieve\,-- SPLADE-only}    & $0.269$ & $0.36$  \\
            \textbf{w/ \extract\,-- keys-only}    & $0.138$ & $0.23$  \\
        \bottomrule
    \end{tabular} 
    \vspace*{-0.3cm}
    \caption{Ablation studies on the \benchmark \textbf{\textit{dev}} set.}
    \label{tab:ablation}
    \vspace*{-0.2cm}
}
\end{table}

\vspace{0.1cm}
\myparagraph{\method adapts to low-resource computation}  
Our goal is  a method that can run on user devices.
However, there is high variance in the available compute-power and memory on client hardware, and QUD alone may use cloud services without violating data protection.
Hence, we trained models used in \method of varying sizes:
from $135$M-$3$B parameters for the QUD model,
$4$M-$33$M for the cross-encoders of \retrieve,
and $16$M-$139$M for the \extract model.
Table~\ref{tab:model-sizes} shows the results.

While there is a substantial performance gap between the smallest variant (QUD: XS, Operators: XS)
and the largest variant (L, L), we observe interesting trade-offs 
and compromises.
When the memory footprint has to be low, but computational costs and run-time are secondary,
the (XS, L) variant is a good choice, achieving a Rlx-Hit@1 of $51$\%.
Recall that QUD runs on the
question only, 
without access to user data.
When questions alone are not sensitive, a larger cloud model can be used.


\begin{table} [t]
    \center
    \newcolumntype{H}{>{\setbox0=\hbox\bgroup}c<{\egroup}@{}}
    \newcolumntype{G}{>{\columncolor [gray] {0.90}}c}
    \resizebox*{\columnwidth}{!}{
    \small
    \begin{tabular}{l G c c c c}
        \toprule
             & \textbf{QUD $\rightarrow$} & \textbf{XS} & \textbf{S} & \textbf{M} & \textbf{L} \\
             & \textbf{Operators $\downarrow$} & ($135$M) & ($360$M)  &  ($1$B) & ($3$B) \\
        \midrule
            \multirow{2}{*}{\textbf{XS}} &  {\retrieve} (4M)
                & $0.167$ 
                & $0.185$ 
                & $0.193$ 
                & $0.193$ \\ 
            & {\& \extract} (16M)
                & \relaxed{0.28} 
                & \relaxed{0.28} 
                & \relaxed{0.31} 
                & \relaxed{0.30} \\ 
        \midrule
            \multirow{2}{*}{\textbf{S}} &  {\retrieve} (16M) 
                & $0.240$ 
                & $0.287$ 
                & $0.302$ 
                & $0.302$ \\ 
            & {\& \extract} (31M) 
                & \relaxed{0.37} 
                & \relaxed{0.41} 
                & \relaxed{0.42} 
                & \relaxed{0.42} \\ 
        \midrule
            \multirow{2}{*}{\textbf{M}} &  {\retrieve} (23M) 
                & $0.331$ 
                & $0.353$ 
                & $0.378$ 
                & $0.389$ \\ 
           & {\& \extract} (70M)
                & \relaxed{0.48} 
                & \relaxed{0.51} 
                & \relaxed{0.50} 
                & \relaxed{0.52} \\ 
        \midrule
            \multirow{2}{*}{\textbf{L}} & {\retrieve} (33M) 
                & $0.356$ 
                & $0.364$ 
                & $0.396$ 
                & ${0.400}$ \\ 
             & {\& \extract} (139M)
                & \relaxed{0.51} 
                & \relaxed{0.52} 
                & \relaxed{0.54} 
                & \relaxed{0.55} \\ 
        \bottomrule
    \end{tabular}
    }
    \vspace*{-0.3cm}
    \caption{Effect of model sizes on performance.
    Metric is Hit@1, and Rlx-Hit@1 in parentheses.}
    \label{tab:model-sizes}
    \vspace*{-0.2cm}
\end{table}

\vspace{0.1cm}
\myparagraph{Error analysis}
We 
analyzed
$50$ incorrect answers.
\method failed because of
(i) an incorrect operator tree
(50\%),
(ii) errors by the \retrieve operator
(28\%), or
(iii) errors by the \extract operator
(22\%).
Note that sometimes just a few
false negatives or false positives
can lead to an incorrect answer,
underlining the difficulty of \benchmark.

\vspace*{0.1cm}
Additional results,
including anecdotal examples and run-time analysis, 
are provided in Appendix~\ref{app:results}.

\subsection{Experiment with TimelineQA}

\myparagraph{Setup}
Although TimelineQA
is deemed insufficient for our setting,
we obtained insight on how \method performs on these
questions.
For this purpose, we used their $42$ templates for complex questions,
and adapted them to our realistic user data.
We ran the questions for each of the $6$ test personas in \benchmark.
SFT variants utilize the models trained on \benchmark,
as the limited number of questions
in TimelineQA does not allow for training.

\vspace{0.1cm}
\myparagraph{Results}
Table~\ref{tab:timelineqa} shows the results
on TimelineQA.
\rag shows stronger performance than on the more challenging questions in \benchmark.
\method substantially outperforms both \rag and \cog,
with the small model (SFT) and with GPT4o.


\begin{table} [t]
{
    \small
    \setlength{\tabcolsep}{3pt}
    \renewcommand{\arraystretch}{0.8}
    \centering
    \newcolumntype{H}{>{\setbox0=\hbox\bgroup}c<{\egroup}@{}}
    \newcolumntype{G}{>{\columncolor [gray] {0.90}}c}

    \begin{tabular}{l G c G}
        \toprule
            \textbf{Model $\rightarrow$} & \textbf{GPT4o} & \textbf{LLaMA3.3}  & \textbf{SFT}\\
            \textbf{Method $\downarrow$}  &  ($\gg$100B) &  (70B) & (1B) \\

      \midrule
            \multirow{2}{*}{\textbf{\rag}}  & ${0.321}$ & $\mathbf{0.369}$& $0.214$  \\
             &  \relaxed{0.36}  & \relaxedb{0.41} & \relaxed{0.22} \\

        \midrule
            \multirow{2}{*}{\textbf{\qg}}  & $0.262$ & $0.187$& $0.135$  \\
             &  \relaxed{0.33}  & \relaxed{0.22} & \relaxed{0.17} \\

        \midrule
            \multirow{2}{*}{\textbf{\method} (ours)}    & $\mathbf{0.357}$  & ${0.286}$   & $\mathbf{0.313}$   \\
               & \relaxedb{0.39}  &  \relaxed{0.31}  &  \relaxedb{0.35}  \\
        \bottomrule
    \end{tabular}
    \vspace*{-0.3cm}
    \caption{Hit@1 and Rlx-Hit@1 on TimelineQA.}
    \label{tab:timelineqa}
    \vspace*{-0.2cm}
}
\end{table}





\section{Evaluation via User Study}
\label{sec:user-study}

The \benchmark benchmark is 
designed with synthetic personas
to avoid privacy issues.
The goal of this user study is two-fold:
(i) analyze whether the questions in \benchmark are comparable with real user questions, and
(ii) investigate whether a small-scale variant of \method,
run on end-user devices,
can deal with such realistic information needs.

\subsection{Setup}

We conducted our user study with $20$ undergraduate students.
First, we instructed participants to export their personal data from various
services like Apple, Netflix and Spotify.
Then, users were provided with Docker images
for running \method on their own laptops and exports,
but without giving us access to their personal data.
The Docker images had disabled Internet connectivity, to guarantee the impossibility of data leaks.

We initially instructed each participant to generate $100$ questions,
without restricting them to any question type or sources.
Users then selected a subset of $20$ questions that are answerable based on their personal data,
assuming a hypothetical perfect QA system.
More details 
are given in Appendix~\ref{app:user-study}.

\vspace*{-0.1cm}
\subsection{Results}
\label{sec:user-study-res}

\myparagraph{User and \benchmark questions are similar}
We ran our QUD module on user and benchmark questions,
and compared the structure of the resulting operator trees, abstracting away the individual constants.
For $376$ out of $400$ (=$20$ users$\times$$20$ questions) 
user questions ($94$\%), we were able to map them to an isomorphic  operator tree in \benchmark.
This indicates that the questions in \benchmark indeed capture the complexities of real-life information needs.

\vspace{0.1cm}
\myparagraph{\method performs well in the wild}
For each of their $20$ selected questions, users
ran \method on their own device and data.
We asked them whether the provided answer is
{\em correct}, {\em almost correct} (with minor drifts
tolerated by the user), or 
{\em incorrect}.
By design,
\method
allows
users
to 
identify which
events
were used for answering,
aiding them to understand the correctness of answers.
Users judged ca. $28$\% of the answers 
for their own questions
as fully correct, and $41$\% as almost correct.

Users also evaluated
$20$ \benchmark questions
against their personal data,
with allowance for minor adjustments
to adapt them to their data.
In this setting,
\method achieves even better performance:
$45$\% for Hit@1 (and $60$\% for Rlx-Hit@).

\vspace*{-0.1cm}
\section{Related Work}
\label{sec:related}

\noindent {\bf Personal data management.}
This theme
was initially motivated by the tasks
of desktop search 
\cite{DBLP:conf/cidr/DongH05}
and mail/document search  \cite{DBLP:conf/sigir/DumaisCCJSR03}.
With the increasing abundance of personal data on mobile phones and cloud storage, the topic has been revived \cite{halevy2024personal}.
Recent works target simple retrieval over personal data \cite{ribeiro2022lifelog,DBLP:conf/mir/Gurrin0NHLZRTHB23}.
The closest work to ours is TimelineQA \cite{DBLP:conf/acl/TanD0MSYH23}, which 
conducts experiments with standard \rag and \cog approaches
-- we show that \method substantially outperforms these state-of-the-art baselines.

Orthogonally, there are concerns about platform providers 
``knowing everything'' about users, and the need for humans to stay in full control of their personal data \cite{DBLP:conf/www/MansourSHZCGAB16,DBLP:conf/cikm/KalokyriBM18}.
Our approach follows this rationale.

\vspace{0.1cm}
\noindent {\bf Data verbalization.}
A popular way of using LLMs for multi-source QA
is to verbalize all data and treat it uniformly as if were a collection of natural language sentences \cite{DBLP:conf/naacl/OguzCKPOSGMY22}.
This approach has been pursued for the case of text+tables
\cite{DBLP:conf/acl/YinNYR20,badaro2023transformers}, and for combining knowledge graphs with other sources \cite{christmann2022conversational,christmann2023explainable,DBLP:journals/ws/PramanikARW24,DBLP:conf/acl/ZhangSGXLL24}.
%


\vspace{0.1cm}
\noindent {\bf Question-to-query translation.}
State-of-the-art methods, often referred to as Text2SQL or NL2SQL 
\cite{DBLP:journals/vldb/KatsogiannisMeimarakisK23,DBLP:journals/corr/abs-2408-05109},
leverage LLM pre-training with code repositories (e.g., GitHub) and
fine-tune the model with question-query pairs
\cite{DBLP:conf/nips/PourrezaR23,fan2024combining,DBLP:journals/pvldb/LiLCLT24,DBLP:journals/pvldb/GaoWLSQDZ24,DBLP:conf/sigir/GongS24}.
Importantly, these methods require {\em schema} information about
the tables, column headers and column types of the underlying sources.
For complex schemas or when only partial information 
is provided, the performance drops sharply. 


\vspace{0.1cm}
\noindent {\bf Question decomposition.}
By decomposing a user question, \method aims to understand the user's intent and
derive a suitable representation.
This has been pursued for a variety
of specialized settings, including
\citet{DBLP:conf/www/JiaCW24} for temporal questions,
and \citet{DBLP:conf/edbt/0002CP24} and \citet{liu2024suql} for
delegating sub-tasks to an LLM.
Also, enhanced prompting techniques
like chain-of-thought \cite{wei2022chain}
and tree-of-thoughts \cite{yao2023tree}
have been proposed.
While these approaches enable LLMs to decompose tasks of certain kinds, they
exhibit similar drawbacks as \rag, and do not 
generalize to QA over heterogeneous data sources.
\section{Conclusion}
\label{sec:conclusion}

This paper introduced a new approach,
\method,
for QA over heterogeneous personal data, spanning structured tables and unstructured text.  \method recursively decomposes a user question into an operator tree, which can be efficiently executed on user devices.
In our experiments,
\method outperforms methods based on verbalization (\rag)
or translation (\cog),
and our small-model configurations achieve competitive performance, thus reconciling
the goals of expressive QA and light-weight footprint.

\clearpage

\clearpage\newpage

\section{Limitations}
\label{sec:limitations}

Our experimental evaluation is based on two benchmarks and a small-scale user study.
Obviously, it would be desirable to consider additional datasets and scale up the user study. Neither of these is easily feasible.
Collecting data and questions from real users is highly sensitive.
Even with anonymization techniques, adversaries that have background knowledge (e.g., co-workers of participating users) may obtain privacy-critical information.
This is why we did not pursue this approach.

On the technical side, the \method system supports a limited set of data repositories (calendar, mail, music streaming etc.). 
Extending this would be straightforward as long as a service can cast its data into our data model of events with key-value pairs.
A more important restriction is the system's assumption that the data is materialized, via data exports from the user's accounts.
This is non-critical with synthetic personas or in a controlled user study,
but would be a concern on behalf of real users. 
Ideally, \method would tap into the user's data only through APIs of the underlying services,
giving rise to QA over a federation of 
autonomous and encapsulated services.
While this is unrealistic with the current setup of mobile phones,
pursuing such a federated architecture would be important for future research on reconciling user convenience and data protection.

\section{Ethical Considerations}
\label{sec:ethics}


Our benchmark construction intentionally avoided gathering data from real users, as this would be extremely difficult to properly anonymize.
For adversaries with background knowledge about a user,
even anonymized data would not perfectly prevent privacy breaches.

The user study that we carried out with a small cohort had all participants sign an explicit consent form, and has been approved by the Ethics Review Board.
Moreover and most importantly, it was carried out in a way where the users had full control over each and every step.
All their data was exported from their service without leaving their personal devices.
The software for running \method was packaged into Docker images,
with Internet connection switched off.
Only the user-created questions and relevance labels for answers were given to us,
and the users had the option of removing questions, or removing names by placeholders,
if deemed too sensitive.

\noindent All code and data of this work
is
publicly accessible at
\href{https://reqap.mpi-inf.mpg.de}{https://reqap.mpi-inf.mpg.de}
for reproducibility
and support of follow-up research.

\section*{Acknowledgements}
We would like to thank Rishiraj Saha Roy, Sneha Singhania, and Magdalena Kaiser for their helpful feedback on 
this paper.
We also thank the anonymous reviewers for their constructive comments.

\bibliography{references}

\clearpage\newpage
\appendix
\section*{Appendices}
\section{Details on the \method Methodology}

\subsection{Question Understanding\&Decomposition}
\label{app:qu-training}

\begin{figure*}[t]
\begin{tcolorbox}[colback=gray!20, colframe=black, rounded corners, left=0pt, right=0pt, top=0pt, bottom=0pt]
    \lstset{basicstyle=\tiny\ttfamily, breaklines=true, frame=none}
    
    \begin{lstlisting}
You are given a question and need to generate a plan to infer the answer.
This plan follows Python syntax. You can make use of the following functions:
`QUD (question: str)`
    Used to make a recursive call to the QUD module.

`RETRIEVE (query: str) -> QUExecutionResult`
    Retrieves all events that match the query.
`EXTRACT(l: QUExecutionResult, attr_names: List[str], attr_types: List) -> QUExecutionResult`
    Extracts the values for the provided attributes from each event.
`GROUP_BY(l: QUExecutionResult, attr_names: List[str]) -> QUExecutionResult`
    Function to group the results by the provided attributes.
`JOIN(l1: QUExecutionResult, l2: QUExecutionResult, condition: str) -> QUExecutionResult`
    Function to join two results based on the condition. Condition is a string, and makes use of i1 and i2 to refer
    to items in the list.
`APPLY(l: QUExecutionResult, fct: Callable) -> QUExecutionResult`
    Apply a callable on the list of events in l.
`MAP(l: QUExecutionResult, fct: Callable, res_name: Optional[str]="map_result") -> QUExecutionResult`
    Map a callable to each event or group (GROUP_BY result) in l, and store each result in the key `res_name`.
`FILTER(l: QUExecutionResult, filter: Callable) -> QUExecutionResult`
    Filter the events using the provided callable.
`UNNEST(l: QUExecutionResult, nested_attr_name: str, unnested_attr_name: str) -> QUExecutionResult`
    Derives a single event for each value in the list in key `nested_attr_name`, with the value being stored in 
    `unnested_attr_name`.

`ARGMIN / ARGMAX (l: QUExecutionResult, arg_attr_name: str, val_attr_name: Optional[str]) -> QUExecutionResult`
    Computes the argmin/argmax based on the provided attribute (`arg_attr_name`), and returns the corresponding
    value (`val_attr_name`).
`SUM / AVG / MAX / MIN (l: QUExecutionResult, attr_name: str) -> QUExecutionResult`
    Computes the sum/avg/max/min based on the provided attribute (`attr_name`).

`QUExecutionResult` is a special class that is used to store the results of functions.

Whenever accessing an attribute in a function call, make sure that this attribute is derived in a subsequent call via an
appropriate call (e.g., EXTRACT or MAP). For example, when val_attr_name="workout_type" is used in a ARGMAX call, there has
to be a EXTRACT call with attr_names=["workout_type",...] in the following.

Please provide only the next step, including the recursive QUD call to take (if needed).
Learn from the examples below.
    \end{lstlisting}
    \end{tcolorbox}

    \vspace*{-0.3cm}
    \caption{LLM instruction for generating operator trees. This is followed by $8$ in-context learning examples, which are dynamically chosen per step. Sample ICL examples are shown in Table~\ref{tab:icl-examples}.}
    \label{fig:QUD-instructions}
\end{figure*}

\noindent Figure~\ref{fig:QUD-instructions} shows
the instruction used for the LLM-version of the QUD component.
The in-context learning (ICL) samples are
judiciously crafted to bring out key difficulties
that the model has to deal with, while covering
all OTX operators.
We handcraft a total of $40$ such ICL examples,
based on a diverse set of $40$ questions.
Table~\ref{tab:icl-examples} shows $5$ such examples.
Each ICL example consists of multiple user and assistant turns,
enabling the decomposition of questions.

During LLM-inference for a given sub-question,
we retain the previous input sub-questions and output operators
for context.

\subsection{\method Operators}
\label{app:operators}

This section provides additional detail on the operators
currently in-use within \method.

\vspace{0.1cm}
\myparagraph{\operator{JOIN}}
The \operator{Join} operator has the following inputs and outputs:
\squishlist
\item {\bf Input:} The input comprises {\em two lists of events},
computed as intermediate results for two sub-questions by the
upstream operators (descendants in the tree), and a 
{\em join condition}: a comparison between key-value pairs of the two input lists.
\item {\bf Output:} The result is a set of {\em combined events} (the two sets of key-value pairs are merged)
that contains all event pairs from the two input lists that satisfy the join condition.
\squishend

\noindent Unlike joins in mainstream multi-hop QA, where all joins are equi-joins (often just link traversal in graphs), 
our setting requires support for non-equi joins, most notably, for temporal comparisons with before/after and overlap of time intervals.

Our method for this general join processing
is a sort-merge algorithm.
The two sets of events to be joined
are sorted by the keys that the
join predicate refers to, and the join result is obtained by scanning and merging the two ordered streams.


\vspace{0.1cm}
\myparagraph{\operator{GROUP\_BY}}
The {\operator{Group\_by} operator} has the following inputs and outputs:
\squishlist
\item {\bf Input:} a list of events and a list of keys in this data, by which the events are grouped.
\item {\bf Output:} a partitioning of the events, represented as a list of lists. Each partition (i.e., one of the inner lists) consists of all events that share the same values for the input keys.
\squishend
%
For the implementation, we adopted a hash-based algorithm that maps events into a hash table by their values of the grouping keys.
The simplest case is a single key for grouping, but grouping by multiple keys can often be required as well:
an example is for questions like \phrase{in which month did I run the most?},
which logically require grouping by month and year.

\myparagraph{\operator{FILTER}}
The {\operator{Filter} operator} has the following inputs and outputs:
\squishlist
\item {\bf Input:} a set of events and a filter condition
which is implemented as a function
that can refer to key-value pairs (e.g., \struct{cuisine}=\phrase{Italian}).
\item {\bf Output:} all events that satisfy the filter condition.
\squishend
The implementation supports arbitrary filter conditions expressed as Python code,
including exact matching, numeric matches, temporal comparisons, or sub-string matches.
Within the \operator{Filter} operator, calls to the QUD module are allowed as well,
enabling a flexible framework for various situations.

\vspace{0.1cm}
\myparagraph{\operator{MAP}}
The {\operator{Map} operator} has the following inputs and outputs:
\squishlist
\item {\bf Input:} 
a list of events and a Python function.
\item {\bf Output:} a list of return values of the function applied to each element in the list (e.g., the weekday for each event).
\squishend
This operator is key to plugging in arbitrary functions (similar to user-defined functions (UDFs) in SQL) into the execution plan.
This is frequently needed, for example, for calendar-centric calculations, such as the weekday of an event or the duration of the period between dates.
By enabling such plug-in functions, the \method system is easily extensible.

\vspace{0.1cm}
\myparagraph{\operator{APPLY}}
The {\operator{Apply} operator} has the following inputs and outputs:
\squishlist
\item {\bf Input:} a list of events and a Python function.
\item {\bf Output:} 
a scalar value computed over the entire list.
\squishend
One role of \operator{Apply} is to compute aggregation functions, such as the length of a list (for counting), summing up numeric values, etc.

\vspace{0.1cm}
\myparagraph{\operator{UNNEST}}
The SQL-inspired {\operator{Unnest} operator} has the following inputs and outputs:
\squishlist
\item {\bf Input:} a list of events, and a key pointing to the nested lists.
\item {\bf Output:} an extended list of events, which holds an independent event for each item in the nested lists, flattening this key-value pair.
\squishend

An example use-case of \operator{Unnest} is
a user question like \phrase{Artist I listened to the most?},
that requires grouping by individual artist names,
while the events themselves have a key \struct{artists}
holding multiple names.
\operator{Unnest} creates an independent event for each artist name in the list,
enabling to answer such questions correctly.

The implementation of \operator{Apply}, \operator{Map} and \operator{Unnest} is straightforward.
The remaining operators
(\operator{Argmin}, \operator{Argmax}, \operator{Min}, \operator{Max}, \operator{Sum}, and \operator{Avg}) are self-explanatory.

\section{Details on the \benchmark Benchmark}
\label{app:benchmark}

\subsection{Details on the \benchmark Construction}
\label{app:benchmark-construction}

\myparagraph{Persona details}
For each fictional persona, the following information is collected:
\textit{name}, \textit{gender}, \textit{birth date/city}, \textit{mother}, \textit{father}, \textit{siblings}, \textit{kids} (incl. birth date), \textit{pets} (incl. type, start/end date), \textit{friends}, \textit{education} (incl. city, start/end date),
\textit{career} (incl. company, city, start/end date), \textit{residences} (incl. start/end date).
Further, we collect user preferences:
\textit{favorite music genres}, \textit{movie genres}, \textit{TV series genres}, \textit{shopping categories}, \textit{travel regions}, \textit{cuisines}, \textit{hobbies} and \textit{workouts}, as well as favorite instances (e.g., favorite songs)
and the average frequency (e.g., average movies per week).

\vspace{0.1cm}
\myparagraph{Canonicalized event types}
\benchmark integrates the following canonicalized event types:
\textit{anniversaries} (birthday, child's birthday, partner anniversary, valentines day, halloween, christmas, New Year's Eve, pet anniversary), \textit{doctor appointments} (dentist, gp, ophthalmologist, dermatologist, paediatrician, veterinarian), \textit{personal milestones} (new job, start of new education, new residence, marriage, new kid, new pet), \textit{traveling} (incl. 2-5 notable events), \textit{meetings} (with friends and family), \textit{streaming} (music, TV series, movies), \textit{online shopping} and \textit{workouts}.

\vspace{0.1cm}
\myparagraph{Natural user behavior}
A key criterion when creating user data is naturalness.
Hence, we aim to derive sequences of events that are as natural as possible.
Consider the music streaming behavior as an example:
it is very unlikely that someone listens to a single song at 2AM and then to another song at 5AM.
We thus base the sequences of events on user behavior patterns.
Such user behavior patterns are derived from streaming data of the authors
and colleagues (only the patterns, no critical data is used).
These patterns are then used to define the sequence of events
for streaming data, workouts and online shopping.

\vspace{0.1cm}
\myparagraph{Cleaning of unnatural questions}
The generation process may still yield some questions that are misinformed for specific personas.
For example,
there could be question templates asking for the birth date of the first-born child,
but not all personas have children.
Such {\em empty-answer} questions
(no output, or a count of 0 for aggregations), are unnatural
which is why we removed them for the respective personas.

\vspace*{0.1cm}
\myparagraph{Quality check}
Data quality was essential for this work. To this end, we performed the user study (Section~\ref{sec:user-study}) for exactly this purpose:
comparing the benchmark data and questions against those from real users, obtaining additional insights from real users, and this way re-assessing the significance of our main results.
Section~\ref{sec:user-study-res}, first paragraph, compares the structure of questions from these two settings: synthetic vs. real users.
We observe that a very large fraction of the user’s questions ($94$\%) have structurally isomorphic counterparts in the \benchmark dataset. Obviously, the literals between these cases are not comparable (e.g., real users’ friends would have names different from the ones in \benchmark).
But this comparison generally shows that the real-user questions are structurally very similar to those in \benchmark.

Apart from the results derived from the user study, we annotated $100$ randomly sampled questions in \benchmark.
We found that $98$\% of questions are relevant.
We further annotated the ground-truth answers for these 100 sample questions. We observed that $97$\% of the answers are correct,
with minor issues otherwise.
Finally, we sampled 100 user events in \benchmark, and investigated whether the events are realistic user events, and whether the LLM-verbalizations (mails, calendar entries and social media posts) are semantically equivalent. We found that 98\% of events do indeed express realistic user events,
and LLM-verbalizations are semantically equivalent in all cases.

\vspace*{0.1cm}
\myparagraph{Usage of LLaMA-70B}
The version originally submitted to ACL rolling review 
was using LLaMA-8B\footnote{\href{https://huggingface.co/meta-llama/Meta-Llama-3-8B-Instruct}{meta-llama/Meta-Llama-3-8B-Instruct}} 
for verbalizing the canonicalized events in \benchmark.
For enhanced diversity, and even more realistic data,
we utilized LLaMA-70B\footnote{\href{https://huggingface.co/meta-llama/Meta-Llama-3-70B-Instruct}{meta-llama/Meta-Llama-3-70B-Instruct}} instead in the revised version of this work.
This explains slight derivations of results from the
submitted version of the paper --
generally, through the more diverse
data, results of \method improved.

\vspace{0.1cm}
\myparagraph{License}
The \benchmark dataset is released with a \href{https://creativecommons.org/licenses/by/4.0/}{CC BY 4.0} license to enable others to work with our data and advance the state-of-the-art.

\vspace{0.1cm}
\myparagraph{Data demographics}
All data in \benchmark is English.
The personas collected are artificial, and do not represent real human beings.

\subsection{\benchmark Examples}

\myparagraph{Example questions}
Table~\ref{tab:perqa-questions} shows example questions
from our new \benchmark dataset,
demonstrating challenges such as aggregation,
temporal conditions,
numerical constraints,
or joining information from different sources.

\vspace*{0.1cm}
\myparagraph{Example user data}
Table~\ref{tab:perqa-data} illustrates example events in \benchmark.
Such events can be of an unstructured, semi-structured or structured nature,
and form the
personal data in \benchmark
that systems have to process for answering the questions.


\begin{figure*}[t]
\begin{tcolorbox}[colback=gray!20, colframe=black, rounded corners, left=0pt, right=0pt, top=0pt, bottom=0pt]
    \lstset{basicstyle=\tiny\ttfamily, breaklines=true, frame=none}
    
    \begin{lstlisting}
You are an assistant that generates questions, and the corresponding SQL query for deriving the answer, on personal data.

You are provided with the SQL schema of the underlying DB, and a set of example questions and SQL queries.
Your task is to generate 25 questions and their SQL queries.

Make sure that the questions are natural, and the SQL query is compliant with DuckDB -- make use of the provided SQL schema
and examples. Please generate complex but realistic questions and SQL queries -- based on multiple tables and columns, with
joins, group by's, and variable declarations.

Also try to generate diverse questions: formulations can be ad-hoc or well-formed, and their sentence structure could
differ. Remember that question should be natural: do not spell out the exact table names or columns, but make it realistic.

Your output should be a plain JSON object, with no other text.
    \end{lstlisting}
    \end{tcolorbox}

    \vspace*{-0.3cm}
    \caption{LLM instruction for generating complex questions. In addition, we randomly sample $20$ questions that are provided as ICL examples.}
    \label{fig:prompt-questions}
\end{figure*}

\begin{table}
    \footnotesize
    \newcolumntype{G}{>{\columncolor [gray] {0.95}}l}
    \resizebox*{\columnwidth}{!}{
    \begin{tabular}{p{9cm}}
        \toprule
            \multicolumn{1}{G}{\textbf{Questions -- Workouts}}\\
        \midrule
            - \phrase{What is the total distance I ran in kilometers since I started working at Shopify?}\\
            - \phrase{My maximum heart rate during the most recent workout?}\\
            - \phrase{How many km did I run in the first week of 2019?}\\
            - \phrase{How many workouts did I do in 2018 with a maximum heart rate of more than 170?}\\

        \midrule
            \multicolumn{1}{G}{\textbf{Questions -- Music}}\\
        \midrule
            - \phrase{Which song did I listen to the most in the last month?}\\
            - \phrase{What is the name of the song I listened to the most in the summer of 2022?}\\
            - \phrase{Which song did I listen to most after a badminton workout?}\\
            - \phrase{which artist did I listen to most?}\\

        \midrule
            \multicolumn{1}{G}{\textbf{Questions -- Movies / TV series}}\\
        \midrule
            - \phrase{how many different movies did I watch?}\\
            - \phrase{the TV series I watched the most in summer 2022?}\\
            - \phrase{What's the total number of hours spent on TV series?}\\
            - \phrase{Which day of the week do I stream TV series the most?}\\

        \midrule
            \multicolumn{1}{G}{\textbf{Questions -- Mail / social media }}\\
        \midrule
            - \phrase{Who was the recipient of the first email I sent this year?}\\
            - \phrase{Who are my top 3 most contacted people via email?}\\
            - \phrase{How many emails did I send after 6 PM?}\\
            - \phrase{What was the last social media post I made?}\\

        \midrule
            \multicolumn{1}{G}{\textbf{Questions -- Others}}\\
        \midrule
            - \phrase{How many times did I meet with Carla Díaz at a restaurant?}\\
            - \phrase{What day of the week do I most frequently schedule doctor appointments?}\\
            - \phrase{What was the total amount spent on online purchases last year?}\\
            - \phrase{How many times did I meet at a park in 2022?}\\

        \bottomrule

    \end{tabular}
    }
    \vspace*{-0.2cm}
    \caption{Example questions in \benchmark.}
    \label{tab:perqa-questions}
\end{table}

\begin{table*} [t]
    \small
    \centering
    \newcolumntype{G}{>{\columncolor [gray] {0.90}}c}
    \begin{tabular}{l c c c}
        \toprule
            \textbf{Module} & \textbf{Size} & \textbf{\#Parameters} & \textbf{Model} \\
        \midrule
            \multirow{4}{*}{\textbf{QUD}} & XS & $135$M & \href{https://huggingface.co/HuggingFaceTB/SmolLM2-135M-Instruct}{HuggingFaceTB/SmolLM2-135M-Instruct} \\
            & S & $360$M & \href{https://huggingface.co/HuggingFaceTB/SmolLM2-360M-Instruct}{HuggingFaceTB/SmolLM2-360M-Instruct} \\
            & M & $1$B & \href{https://huggingface.co/meta-llama/Llama-3.2-1B-Instruct}{meta-llama/Llama-3.2-1B-Instruct} \\
            & L & $3$B & \href{https://huggingface.co/meta-llama/Llama-3.2-3B-Instruct}{meta-llama/Llama-3.2-3B-Instruct} \\
        \midrule
            \multirow{4}{*}{\textbf{\retrieve}} & XS & $4$M & \href{https://huggingface.co/cross-encoder/ms-marco-TinyBERT-L-2-v2}{cross-encoder/ms-marco-TinyBERT-L-2-v2} \\
            & S & $16$M & \href{https://huggingface.co/cross-encoder/ms-marco-MiniLM-L-2-v2}{cross-encoder/ms-marco-MiniLM-L-2-v2} \\
            & M & $23$M & \href{https://huggingface.co/cross-encoder/ms-marco-MiniLM-L-6-v2}{cross-encoder/ms-marco-MiniLM-L-6-v2} \\
            & L & $33$M & \href{https://huggingface.co/cross-encoder/ms-marco-MiniLM-L-12-v2}{cross-encoder/ms-marco-MiniLM-L-12-v2} \\
        \midrule
            \multirow{4}{*}{\textbf{\extract}} & XS & $16$M & \href{https://huggingface.co/google/t5-efficient-tiny}{google/t5-efficient-tiny} \\
            & S & $31$M & \href{https://huggingface.co/google/t5-efficient-mini}{google/t5-efficient-mini} \\
            & M & $70$M & \href{https://huggingface.co/lucadiliello/bart-small}{lucadiliello/bart-small} \\
            & L & $139$M & \href{https://huggingface.co/facebook/bart-base}{facebook/bart-base} \\
        \bottomrule
    \end{tabular} 
    \vspace*{-0.3cm}
    \caption{Models used for the model-size analysis in Table~\ref{tab:model-sizes}.}
    \label{tab:cfg-model-sizes}
\end{table*}

\section{Details on Configuration}
\label{app:setup}

\myparagraph{Main results}
We fine-tuned the QUD model for $4$ epochs.
The learning rate was tuned on a with-held dev set, and was eventually set to $5*10^{-6}$,
with a linear scheduling. The warm-up ratio was set to $0.05$, with a weight decay of $0.01$,
and a batch size of $4$.
Our train set for the QUD model had $17{,}933$ instances, and the dev set $1{,}135$.
These were obtained by running the GPT4o version of \method on the train and dev sets, sampling $n$=$3$ generations per input with a temperature of $0.1$.
This results in multiple candidate operator trees per training instance -- 
we retain only the ones leading to a correct answer, and a maximum of $3$ per instance.

The \retrieve models (size L) were fine-tuned for $10$ epochs with a learning rate of $5*10^{-5}$,
a batch size of $64$, and a weight decay of $0.001$.
The \extract model (size L) was fine-tuned for $3$ epochs with a learning rate of $1*10^{-4}$,
with a batch size of $16$, and a weight decay of $0.01$.
All our models were trained using an epoch-wise evaluation strategy,
and with AdamW as optimizer.

\method is implemented in Python, and the generated operators thus follow Python syntax as well.

\vspace{0.1cm}
\myparagraph{Baselines}
Baselines generally follow the same training scheme.
For the \rag baseline, we tuned
the number of events in the input on
the dev set, and found the performance to be best with $100$ events (with GPT4o and LLaMA3.3) and $10$ events (with the SFT variant).
Note that including $>$$500$ events exceeded the context window of GPT4o and LLaMA3.3.

\vspace{0.1cm}
\myparagraph{Resources}
We trained all our models on a single NVIDIA Tesla H100NVL GPU.
Training of the QUD model took up to $2.5$ hours (for the $3$B LLaMA model),
for the \retrieve models up to $1.5$ hours (for the $33$M parameter model),
and for the \extract model up to $1.5$ hours as well (for the $139$M model).
For inference of the $70$B LLaMA models,
we used $4$ NVIDIA Tesla H100NVL GPUs.

\vspace{0.1cm}
\myparagraph{Model-size analysis}
Table~\ref{tab:cfg-model-sizes} lists the details on the models used in
the experiment in Table~\ref{tab:model-sizes},
that compares \method with models of varying sizes.

\vspace{0.1cm}
\myparagraph{Licenses}
Our \method code is released with an MIT license to enable
others to re-use components and further advance the state-of-the-art.
TimelineQA was used in experiments under a \href{https://creativecommons.org/licenses/by-nc/4.0/}{CC BY-NC 4.0} license.

\begin{table*} [t]
    \small
    \centering
    \newcolumntype{G}{>{\columncolor [gray] {0.90}}c}
    \resizebox*{\textwidth}{!}{
    \begin{tabular}{l G G G c c c G G G}
        \toprule
            \textbf{Model / Persona $\rightarrow$} & \multicolumn{3}{G}{\textbf{Persona 1} ($489$ questions)} & \multicolumn{3}{c}{\textbf{Persona 2} ($495$ questions)} & \multicolumn{3}{G}{\textbf{Persona 3} ($530$ questions)} \\
            \textbf{Method $\downarrow$} & \textbf{GPT4o} & \textbf{LLaMA3.3}  & \textbf{SFT} & \textbf{GPT4o} & \textbf{LLaMA3.3}  & \textbf{SFT} & \textbf{GPT4o} & \textbf{LLaMA3.3}  & \textbf{SFT} \\
        \midrule
            \multirow{2}{*}{\textbf{\rag}} 
            & $0.180$ & $0.174$ & $0.037$ 
            & $0.154$ & $0.125$ & $0.038$ 
            & $0.147$ & $0.117$ & $0.032$ \\
            & \relaxed{0.26} & \relaxed{0.25} & \relaxed{0.06}
            & \relaxed{0.21} & \relaxed{0.19} & \relaxed{0.06}
            & \relaxed{0.21} & \relaxed{0.17} & \relaxed{0.06} \\
        \midrule
            \multirow{2}{*}{\textbf{\qg}} 
            & $0.344$ & $0.254$ & $0.358$ 
            & $0.349$ & $0.269$ & $0.337$ 
            & $0.326$ & $0.240$ & $0.319$ \\
            & \relaxed{0.45} & \relaxed{0.33} & \relaxed{0.48}
            & \relaxed{0.47} & \relaxed{0.35} & \relaxed{0.48}
            & \relaxed{0.46} & \relaxed{0.33} & \relaxed{0.48} \\
        \midrule
            \multirow{2}{*}{\textbf{\method} (ours)} 
            & $\mathbf{0.421}$ & $\mathbf{0.348}$ & $\mathbf{0.427}$ 
            & $\mathbf{0.402}$ & $\mathbf{0.331}$ & $\mathbf{0.394}$ 
            & $\mathbf{0.425}$ & $\mathbf{0.353}$ & $\mathbf{0.404}$ \\
            & \relaxedb{0.54} & \relaxedb{0.46} & \relaxedb{0.56}
            & \relaxedb{0.54} & \relaxedb{0.47} & \relaxedb{0.53}
            & \relaxedb{0.56} & \relaxedb{0.49} & \relaxedb{0.55} \\
        \bottomrule
    \end{tabular}
    }
    \vspace*{-0.2cm}
    \caption{In-depth results for \benchmark \textbf{\textit{test}} personas 1-3. Metric is Hit@1, and Rlx-Hit@1 in parentheses.}
    \label{tab:per-persona-results1}
    
    \vspace{0.5cm}

    \resizebox*{\textwidth}{!}{
    \begin{tabular}{l G G G c c c G G G}
        \toprule
            \textbf{Model / Persona $\rightarrow$} & \multicolumn{3}{G}{\textbf{Persona 4} ($460$ questions)} & \multicolumn{3}{c}{\textbf{Persona 5} ($524$ questions)} & \multicolumn{3}{G}{\textbf{Persona 6} ($546$ questions)} \\
            \textbf{Method $\downarrow$} & \textbf{GPT4o} & \textbf{LLaMA3.3}  & \textbf{SFT} & \textbf{GPT4o} & \textbf{LLaMA3.3}  & \textbf{SFT} & \textbf{GPT4o} & \textbf{LLaMA3.3}  & \textbf{SFT} \\
        \midrule
            \multirow{2}{*}{\textbf{\rag}} 
            & $0.133$ & $0.089$ & $0.015$ 
            & $0.166$ & $0.130$ & $0.029$ 
            & $0.119$ & $0.103$ & $0.024$ \\
            & \relaxed{0.17} & \relaxed{0.15} & \relaxed{0.08}
            & \relaxed{0.22} & \relaxed{0.17} & \relaxed{0.06}
            & \relaxed{0.17} & \relaxed{0.16} & \relaxed{0.04} \\
        \midrule
            \multirow{2}{*}{\textbf{\qg}} 
            & $0.272$ & $0.209$ & $0.285$ 
            & $0.307$ & $0.227$ & $0.292$ 
            & $0.315$ & $0.234$ & $0.299$ \\
            & \relaxed{0.41} & \relaxed{0.31} & \relaxed{0.46}
            & \relaxed{0.44} & \relaxed{0.32} & \relaxed{0.45}
            & \relaxed{0.45} & \relaxed{0.33} & \relaxed{0.46} \\
        \midrule
            \multirow{2}{*}{\textbf{\method} (ours)} 
            & $\mathbf{0.341}$ & $\mathbf{0.300}$ & $\mathbf{0.357}$ 
            & $\mathbf{0.385}$ & $\mathbf{0.313}$ & $\mathbf{0.368}$ 
            & $\mathbf{0.342}$ & $\mathbf{0.286}$ & $\mathbf{0.335}$ \\
            & \relaxedb{0.49} & \relaxedb{0.45} & \relaxedb{0.52}
            & \relaxedb{0.53} & \relaxedb{0.45} & \relaxedb{0.52}
            & \relaxedb{0.49} & \relaxedb{0.43} & \relaxedb{0.50} \\
        \bottomrule
    \end{tabular}

    }
    \vspace*{-0.2cm}
    \caption{In-depth results for \benchmark \textbf{\textit{test}} personas 4-6. Metric is Hit@1, and Rlx-Hit@1 in parentheses.}
    \label{tab:per-persona-results2}
\end{table*}

\section{Details on Experimental Results}
\label{app:results}

\myparagraph{Results per-persona}
Table~\ref{tab:per-persona-results1} and Table~\ref{tab:per-persona-results2}
provide results on the test set of \benchmark for each persona individually.
All of the methods show variance across the different personas.
Notably, \method performs better than baselines across all personas and models.

\vspace{0.1cm}
\myparagraph{Run-time analysis}
We conducted a run-time analysis for \method, and measured
an average run-time of $0.56$ s per question for the QUD stage (size M), and an average run-time of $11.53$ s for the OTX stage (size L), leading to an average run-time of 
$12.09$ s.
The
median run-time is much lower with
$3.57$ s.
Table~\ref{tab:operator-runtimes} shows the average run-time per operator. Unsurprisingly, the \retrieve and \extract operators,
that are based on LMs and run on thousands of inputs for each question, are the most computationally expensive. 





\begin{table} [t]
    \centering
    {
    \footnotesize
    \newcolumntype{G}{>{\columncolor [gray] {0.90}}c}
    \begin{tabular}{l G c}
        \toprule
            \textbf{Operator} & \textbf{Avg. runtime} \\
        \midrule
            \operator{retrieve} & $2.61$ s \\
            \operator{extract} & $5.93$ s \\
        \midrule
            \operator{join} & $2.29$ s \\
            \operator{group\_by} & $4.82 \times 10^{-3}$ s \\
        \midrule
            \operator{filter} & $1.65 \times 10^{-2}$ s \\
            \operator{map} & $3.44 \times 10^{-1}$ s \\
            \operator{apply} & $4.82 \times 10^{-3}$ s \\
            \operator{unnest} & $3.36 \times 10^{-1}$ s \\
        \midrule
            \operator{argmin} & $3.36 \times 10^{-1}$ s \\
            \operator{argmax} & $8.61 \times 10^{-5}$ s \\
            \operator{min} & $1.74 \times 10^{-4}$ s \\
            \operator{max} & $1.33 \times 10^{-4}$ s \\
            \operator{sum} & $1.20 \times 10^{-4}$ s \\
            \operator{avg} & $1.13 \times 10^{-4}$ s \\
        \bottomrule
    \end{tabular} 
    }
    \vspace*{-0.3cm}
    \caption{Average operator run-times.}
    \label{tab:operator-runtimes}
    \vspace*{-0.4cm}
\end{table}

\vspace{0.1cm}
\myparagraph{Anecdotal examples}
Table~\ref{tab:anecdotes} shows anecdotal examples for operator trees generated by QUD.

\vspace{0.1cm}
\myparagraph{Anecdotal failure cases}
QA over heterogeneous personal data often requires aggregating or ordering information from hundreds of events.
For such questions, even minor errors in one of the operators can lead to an incorrect answer
-- as observed in \benchmark and in the user study -- making the task
a real challenge even with recent advances in NLP.

We report one such anecdote from the user study (we asked the respective user for consent on reporting this anecdote):
for the user query \phrase{most expensive purchase},
\method derived the correct answer among the top few results
(based on the available user data).
However, a mail from a mailing list,
mentioning that an apartment was available for a rent of $730$€,
was incorrectly classified as relevant by \retrieve.
The corresponding value, $730$€ was then found to be the highest,
leading to an incorrect answer.
Note that such a case would yield a Hit@1 
and Rlx-Hit@1 of $0$ on \benchmark.


\begin{table*} [th]
    \footnotesize
    \newcolumntype{G}{>{\columncolor [gray] {0.95}}l}
    \resizebox*{\textwidth}{!}{
    \begin{tabular}{p{\textwidth}}
        \toprule
            \multicolumn{1}{G}{\textbf{Unstructured sources}} \\
        \midrule
            \{\datakey{source}: \phrase{\textbf{mail}}, \datakey{subject}: \phrase{Halloween was a blast!}, \datakey{timestamp}: \phrase{2016-11-01T10:44:21Z}, \datakey{sender}: \phrase{Isabella Ruiz}, \datakey{recipient}: \phrase{Lucia Hernández}, \datakey{text}: \phrase{Hey Lucia, hope you're doing well! I just wanted to share with you that I had a fantastic time celebrating Halloween with Carla last night. We stayed in and had a movie marathon, but it was still super fun. I loved the decorations and the costumes - Carla's was amazing! I'll have to show you the pictures soon. How about we plan something for this weekend? Maybe we can go for a hike or something? Let me know what you think. Take care, Isabella}\} \\
        \midrule
            \{\datakey{source}: \phrase{\textbf{social\_media}}, \datakey{text}: \phrase{Pumped up after a grueling 117-minute weight training session. My heart rate reached a max of 182, but I managed to keep it steady throughout. Feeling strong and energized now! \#weightlifting \#fitnessmotivation}\} \\
        \midrule
            \multicolumn{1}{G}{\textbf{Semi-structured sources}} \\
        \midrule
            \{\datakey{source}: \phrase{\textbf{calendar}}, \datakey{start\_date}: \phrase{2024-08-19}, \datakey{start\_time}: \phrase{12:00:00}, \datakey{end\_date}: \phrase{2024-08-19}, \datakey{end\_time}: \phrase{13:00:00}, \datakey{summary}: \phrase{Lunch with Mum and Dad}, \datakey{location}: \phrase{The Parthenon}, \datakey{description}: \phrase{Greek food}\} \\
        \midrule
            \multicolumn{1}{G}{\textbf{Structured sources}} \\
        \midrule
            \{\datakey{source}: \phrase{\textbf{online\_purchase}}, \dots, \datakey{order}: \phrase{1 x Cosmic Funk}, \datakey{product\_quantity}: \phrase{1}, \datakey{product}: \phrase{Cosmic Funk}, \datakey{category}: \phrase{CDs \& Vinyl}, \datakey{price}: \phrase{5.99 EUR}\} \\
        \midrule
            \{\datakey{source}: \phrase{\textbf{tvseries\_stream}}, \dots, \datakey{duration}: \phrase{2460}, \phrase{stream\_end\_time}: \phrase{00:24:17}, \datakey{stream\_style}: \phrase{tv\_series}, \datakey{tvseries\_title}: \phrase{Scrubs}, \datakey{season\_name}: \phrase{Scrubs, season 1}, \datakey{episode\_name}: \phrase{My Nickname}, \datakey{episode\_number}: \phrase{10}, \datakey{tvseries\_season}: \phrase{1}, \datakey{duration\_unit}: \phrase{seconds}\} \\
        \midrule
            \{\datakey{source}: \phrase{\textbf{movie\_stream}}, \dots, \datakey{duration}: \phrase{6660}, \datakey{stream\_end\_time}: \phrase{01:10:41}, \datakey{stream\_style}: \phrase{movie}, \datakey{movie\_title}: \phrase{Beverly Hills Cop III}, \datakey{stream\_full\_title}: \phrase{Beverly Hills Cop III}, \datakey{duration\_unit}: \phrase{seconds}\} \\
        \midrule
            \{\datakey{source}: \phrase{\textbf{\textit{workout}}}, \dots, \datakey{workout\_type}: \phrase{soccer}, \datakey{duration}: \phrase{126}, \datakey{duration\_unit}: \phrase{min}, \datakey{minimum\_heart\_rate}: \phrase{120}, \datakey{maximum\_heart\_rate}: \phrase{188}, \datakey{average\_heart\_rate}: \phrase{156.87}\} \\
        \bottomrule

    \end{tabular}
    }
    \vspace*{-0.2cm}
    \caption{Example personal data in \benchmark.
    }
    \label{tab:perqa-data}
\end{table*}

\vspace*{-0.1cm}
\section{Details on User Study}
\label{app:user-study}

\myparagraph{Data exports}
Our current implementation supports data exports from Amazon Marketplace, Spotify, Netflix, Apple, Google, as well as mails in the \phrase{.mbox} format and calendar data in the \phrase{.ics} format.
Note that it is trivial to extend
our implementation with additional services, such as streaming data from Amazon Prime. This simply requires providing the data in the format expected by \method, i.e., as sets of key-value pairs.

\vspace*{0.1cm}
\myparagraph{Question creation}
Each participant of our user study created $100$ questions, that capture
their real information needs on their personal data, first.
We provided as little instructions as possible,
to avoid biasing them towards a specific type of questions.
For the evaluation, the participants then chose $20$ questions
that are answerable with their personal data, the sources
we support, and the data in current exports.
For example, the Spotify exports for music streams would not
include the genre or album of the songs, which appeared
in many user questions.
Similarly, we observed questions about watchlists on Netflix,
like {\small \phrase{What is the name of the movies or series that has been in the user Emma's lists for the longest time?}},
that are not covered in current data exports.
All $2{,}005$ questions collected in our user study, including the ones not used for evaluation,
are available at \href{https://reqap.mpi-inf.mpg.de}{https://reqap.mpi-inf.mpg.de}.

\vspace*{0.1cm}
\myparagraph{QUD model}
When conducting the user study,
we were using a preliminary version of the QUD model.
Specifically, we used CodeT5-large~\cite{wang2021codet5} with $738$M parameters,
fine-tuned as outlined in Section~\ref{sec:QUD}\footnote{\href{https://huggingface.co/Salesforce/codet5-large}{Salesforce/codet5-large}}.
We later found variants based on LLaMA-1B to perform substantially better (on \benchmark),
while demonstrating comparable or faster runtime.
Notably, compared to sequence-to-sequence models, decoder-only models can be run
in a quantized manner much more easily.
We hypothesize that the usage of LLaMA-1B might have further improved
the results of the user study
reported in Section~\ref{sec:user-study}.
For the \retrieve operator, we used the model of size S ($16$M), and for the \extract operator we used XS ($16$M).

\vspace{0.1cm}
\myparagraph{Payment}
We hired local
students.
Participants were paid \$$15$ per hour for the user study,
which is well above the local minimum wage.


\begin{table*} [t]
    \footnotesize
    \def\arraystretch{1.1}%
    \newcolumntype{G}{>{\columncolor [gray] {0.95}}l}
    \resizebox*{\textwidth}{!}{
    \begin{tabular}{l}
        \toprule
\multicolumn{1}{G}{\textbf{Anecdotal example 1} -- \phrase{How much money did I spend on online purchases in March 2022?}} \\
        \midrule
        
(0) \qud{QUD("How much money did I spend on online purchases in March 2022?")}\\
(1) SUM(l=\qud{QUD("my online purchases in March 2022 with amounts")}, attr\_name="amount\_spent")\\ 
(2) EXTRACT(l=\qud{QUD("my online purchases in March 2022")}, attr\_names=["amount\_spent"], attr\_types=[float])\\ 
(3) FILTER(l=\qud{QUD("my online purchases with date")},\\ \hspace{0.2cm} filter=lambda attr: attr["purchase\_date"].year == 2022 and attr["purchase\_date"].month == 3)\\ 
(4) EXTRACT(l=\qud{QUD("my online purchases")}, attr\_names=["purchase\_date"], attr\_types=[date.fromisoformat])\\ 
(5) RETRIEVE(query="my online purchases")\\ 

        \midrule
\multicolumn{1}{G}{\textbf{Anecdotal example 2} -- \phrase{First football training after I started as Engineer -- when was it?}} \\
        \midrule
        
(0) \qud{QUD("First football training after I started as Engineer -- when was it?")} \\
(1) MIN(l=\qud{QUD("football training sessions after I started as Engineer")}, attr\_name="start\_datetime")\\ 
(2) FILTER(l=\qud{QUD("my football training sessions with datetime")},\\ \hspace{0.2cm} filter=lambda attr: attr["start\_datetime"] >=\qud{QUD("first start datetime as Engineer")}.result)\\ 
(3.1) EXTRACT(l=\qud{QUD("my football training sessions")}, attr\_names=["start\_datetime"], \\ \hspace{0.2cm}attr\_types=[datetime.fromtimestamp])\\ 
(4.1) RETRIEVE(query="I played football")\\ 
(3.2) MIN(l=\qud{QUD("start datetime as Engineer")}, attr\_name="start\_datetime")\\ 
(4.2) EXTRACT(l=\qud{QUD("I started as Engineer")}, attr\_names=["start\_datetime"],\\ \hspace{0.2cm} attr\_types=[datetime.fromtimestamp])\\ 
(5.2) RETRIEVE(query="I started as Engineer")\\ 

        \midrule
\multicolumn{1}{G}{\textbf{Anecdotal example 3} -- \phrase{which restaurants did we visit when in Bali, Indonesia}} \\
        \midrule
        
(0) \qud{QUD("Which doctor's appointment was the earliest in the day?")} \\
(1) EXTRACT(l=\qud{QUD("restaurants we visited in Bali, Indonesia")}, attr\_names=["restaurant\_name"], \\ \hspace{0.2cm} attr\_types=[str])\\ 
(2) JOIN(l1=\qud{QUD("restaurants we visited with date")}, \\ \hspace{0.2cm} l2=\qud{QUD("we were in Bali, Indonesia with start and end date")}, \\ \hspace{0.2cm} condition="i1.visit\_date >= i2.start\_date and i1.visit\_date <= i2.end\_date")\\ 
(3.1) EXTRACT(l=\qud{QUD("restaurants I visited")}, attr\_names=["visit\_date", "restaurant\_name"], \\ \hspace{0.2cm} attr\_types=[date.fromisoformat, str])\\ 
(4.1) RETRIEVE(query="restaurants I visited")\\ 
(3.2) EXTRACT(l=\qud{QUD("I was in Bali, Indonesia")}, attr\_names=["start\_date", "end\_date"], \\ \hspace{0.2cm} attr\_types=[date.fromisoformat, date.fromisoformat])\\ 
(4.2) RETRIEVE(query="I was in Bali, Indonesia")\\ 

        \midrule
\multicolumn{1}{G}{\textbf{Anecdotal example 4} -- \phrase{Which doctor's appointment was the earliest in the day?}} \\
        \midrule

(0) \qud{QUD("Which doctor's appointment was the earliest in the day?")} \\
(1) ARGMIN(l=\qud{QUD("my doctor's appointments with start time")}, arg\_attr\_name="start\_time", \\ \hspace{0.2cm} val\_attr\_name="appointment\_details")\\ 
(2) EXTRACT(l=\qud{QUD("my doctor's appointments")}, attr\_names=["start\_time", "appointment\_details"], \\ \hspace{0.2cm} attr\_types=[time.fromisoformat, str])\\ 
(3) RETRIEVE(query="my doctor's appointments")\\ 

 \midrule
\multicolumn{1}{G}{\textbf{Anecdotal example 5} -- \phrase{how many products did I buy online in the last 6 months?}} \\
        \midrule
(0) \qud{QUD("how many products did I buy online in the last 6 months?")} \\
(1) SUM(l=\qud{QUD("products bought online in the last 6 months")}, attr\_name="quantity")\\ 
(2) FILTER(l=\qud{QUD("products bought online with purchase date")},\\ \hspace{0.2cm} filter=lambda attr: attr["purchase\_date"] >= (date.today() - relativedelta(months=6)))\\ 
(3) EXTRACT(l=\qud{QUD("products bought online")}, attr\_names=["purchase\_date", "quantity"], \\ \hspace{0.2cm} attr\_types=[date.fromisoformat, int])\\ 
(4) RETRIEVE(query="I bought a product online")\\

        \bottomrule

    \end{tabular}
    }
    \vspace*{-0.2cm}
    \caption{\textbf{Anecdotal examples} of operator trees generated by \method.}
    \label{tab:anecdotes}
\end{table*}

\begin{table*} [t]
\scriptsize
    \newcolumntype{G}{>{\columncolor [gray] {0.95}}l}

\begin{tabularx}{\textwidth}{X}
\toprule

\multicolumn{1}{G}{\textbf{ICL example 1} -- \phrase{On which day did I listen to music the most?}}\\
\midrule
\textbf{User}: Starting with new question.
Input: QUD("On which day did I listen to music the most?")\\
\textbf{Assistant}: ARGMAX(l={{ QUD("number of songs I listened per day?") }}, arg\_attr\_name="num\_songs", val\_attr\_name="start\_date")\\
\textbf{User}: Input: QUD("number of songs I listened per day?")\\
\textbf{Assistant}: MAP(l={{ QUD("my songs I listened to grouped by day") }}, fct=len, res\_name="num\_songs")\\
\textbf{User}: Input: QUD("my songs I listened to grouped by day")\\
\textbf{Assistant}: GROUP\_BY(l={{ QUD("instances I listened to music with date") }}, attr\_names=["start\_date"])\\
\textbf{User}: Input: QUD("instances I listened to music with date")\\
\textbf{Assistant}: EXTRACT(l={{ QUD("I listened to music") }}, attr\_names=["start\_date"], attr\_types=[date.fromisoformat])\\
\textbf{User}: Input: QUD("I listened to music")\\
\textbf{Assistant}: RETRIEVE(query="I listened to music")\\

\midrule
\multicolumn{1}{G}{\textbf{ICL example 2} -- \phrase{how often did I meet with both my parents in the evening?}}\\
\midrule
\textbf{User}: Starting with new question.

Input: QUD("how often did I meet with both my parents in the evening?")\\
\textbf{Assistant}: APPLY(l={{ QUD("I met with both my parents in the evening") }}, fct=len)\\
\textbf{User}: Input: QUD("I met with both my parents in the evening")\\
\textbf{Assistant}: FILTER(l={{ QUD("instances I met with both my parents") }}, filter=lambda attr: attr["start\_time"].hour >= 18 and attr["start\_time"].hour < 24)\\
\textbf{User}: Input: QUD("instances I met with both my parents")\\
\textbf{Assistant}: JOIN(l1={{ QUD("instances I met with my mum") }}, l2={{ QUD("instances I met with my dad") }}, condition="i1.start\_datetime <= i2.end\_datetime and i2.start\_datetime <= i1.end\_datetime")\\
\textbf{User}: Input: QUD("instances I met with my mum")\\
\textbf{Assistant}: RETRIEVE(query="I met with my mum")\\
\textbf{User}: Input: QUD("instances I met with my dad")\\
\textbf{Assistant}: RETRIEVE(query="I met with my dad")\\

\midrule
\multicolumn{1}{G}{\textbf{ICL example 3} -- \phrase{how much money did I spend online the last three years?}}\\
\midrule
\textbf{User}: Starting with new question.

Input: QUD("how much money did I spend online the last three years?")\\
\textbf{Assistant}: SUM(l={{ QUD("my online purchases in the last three years with amounts") }}, attr\_name="amount\_spent")\\
\textbf{User}: Input: QUD("my online purchases in the last three years with amounts")\\
\textbf{Assistant}: EXTRACT(l={{ QUD("my online purchases in the last three years") }}, attr\_names=["amount\_spent"], attr\_types=[float])\\
\textbf{User}: Input: QUD("my online purchases in the last three years")\\
\textbf{Assistant}: FILTER(l={{ QUD("my online purchases with date") }}, filter=lambda attr: attr["purchase\_date"] >= (date.today() - relativedelta(years=3)))\\
\textbf{User}: Input: QUD("my online purchases with date")\\
\textbf{Assistant}: EXTRACT(l={{ QUD("my online purchases") }}, attr\_names=["purchase\_date"], attr\_types=[date.fromisoformat])\\
\textbf{User}: Input: QUD("my online purchases")\\
\textbf{Assistant}: RETRIEVE(query="my online purchases")\\

\midrule
\multicolumn{1}{G}{\textbf{ICL example 4} -- \phrase{which artist did I listen to most when running?}}\\
\midrule
\textbf{User}: Starting with new question.

Input: QUD("which artist did I listen to most when running?")\\
\textbf{Assistant}: ARGMAX(l={{ QUD("number of songs grouped by artist while running") }}, arg\_attr\_name="count", val\_attr\_name="artist")\\
\textbf{User}: Input: QUD("number of songs grouped by artist while running")\\
\textbf{Assistant}: MAP(l={{ QUD("songs grouped by artist while running") }}, fct=len, res\_name="count")\\
\textbf{User}: Input: QUD("songs grouped by artist while running")\\
\textbf{Assistant}: GROUP\_BY(l={{ QUD("songs listened to while running with artist") }}, attr\_names=["artist"])\\
\textbf{User}: Input: QUD("songs listened to while running with artist")\\
\textbf{Assistant}: UNNEST(l={{ QUD("songs listened to while running with artist names") }}, nested\_attr\_name="artist\_names", unnested\_attr\_name="artist")\\
\textbf{User}: Input: QUD("songs listened to while running with artist names")\\
\textbf{Assistant}: EXTRACT(l={{ QUD("songs listened to while running") }}, attr\_names=["artist\_names"], attr\_types=[list])\\
\textbf{User}: Input: QUD("songs listened to while running")\\
\textbf{Assistant}: JOIN(l1={{ QUD("songs I listened to with start and end datetime") }}, l2={{ QUD("I went running with start and end datetime") }}, condition="i1.start\_datetime >= i2.start\_datetime and i1.end\_datetime <= i2.end\_datetime")\\
\textbf{User}: Input: QUD("songs I listened to with start and end datetime")\\
\textbf{Assistant}: EXTRACT(l={{ QUD("songs I listened to") }}, attr\_names=["start\_datetime", "end\_datetime"], attr\_types=[datetime.fromtimestamp, datetime.fromtimestamp])\\
\textbf{User}: Input: QUD("songs I listened to")\\
\textbf{Assistant}: RETRIEVE(query="I listened to a song")\\
\textbf{User}: Input: QUD("I went running with start and end datetime")\\
\textbf{Assistant}: EXTRACT(l={{ QUD("I went running") }}, attr\_names=["start\_datetime", "end\_datetime"], attr\_types=[datetime.fromtimestamp, datetime.fromtimestamp])\\
\textbf{User}: Input: QUD("I went running")\\
\textbf{Assistant}: RETRIEVE(query="I went running")\\

\midrule
\multicolumn{1}{G}{\textbf{ICL example 5} -- \phrase{how often did I meet with Robert in the park?}}\\
\midrule
\textbf{User}: Starting with new question.

Input: QUD("how often did I meet with Robert in the park?")\\
\textbf{Assistant}: APPLY(l={{ QUD("I met with Robert in the park") }}, fct=len)\\
\textbf{User}: Input: QUD("I met with Robert in the park")\\
\textbf{Assistant}: FILTER(l={{ QUD("I met with Robert with location") }}, filter=lambda attr: "park" in attr["location"].lower())\\
\textbf{User}: Input: QUD("I met with Robert with location")\\
\textbf{Assistant}: EXTRACT(l={{ QUD("I met with Robert") }}, attr\_names=["location"], attr\_types=[str])\\
\textbf{User}: Input: QUD("I met with Robert")\\
\textbf{Assistant}: FILTER(l={{ QUD("I met with someone with participants") }}, filter=lambda attr: any("robert" in p.lower() for p in attr["participants"]))\\
\textbf{User}: Input: QUD("I met with someone with participants")\\
\textbf{Assistant}: EXTRACT(l={{ QUD("I met with someone") }}, attr\_names=["participants"], attr\_types=[list])\\
\textbf{User}: Input: QUD("I met with someone")\\
\textbf{Assistant}: RETRIEVE(query="I met with someone")\\

\bottomrule
\end{tabularx}

\vspace{-0.2cm}
\caption{\textbf{ICL examples} used for the LLM-based version of our QUD stage.
We handcrafted $40$ such examples, and retrieve $8$ of them \textit{dynamically}, per step, during inference (via BM25).}
\label{tab:icl-examples}

\end{table*}

\end{document}